\documentclass[runningheads]{llncs}

\usepackage{eccv}

\usepackage{eccvabbrv}

\usepackage{graphicx}
\usepackage{booktabs}

\usepackage[accsupp]{axessibility}  

\usepackage{hyperref}

\usepackage{orcidlink}
\usepackage{multirow}

\usepackage{arydshln}
\usepackage{makecell}
\usepackage{todonotes}
\usepackage{color,soul}
\usepackage{amsfonts}
\usepackage{booktabs}
\usepackage{siunitx}
\usepackage{tabularx}
\usepackage{tcolorbox}
\usepackage[dvipsnames]{xcolor}

\newcommand{\modelname}{Ferret-UI }

\begin{document}

\title{\includegraphics[width=0.7cm]{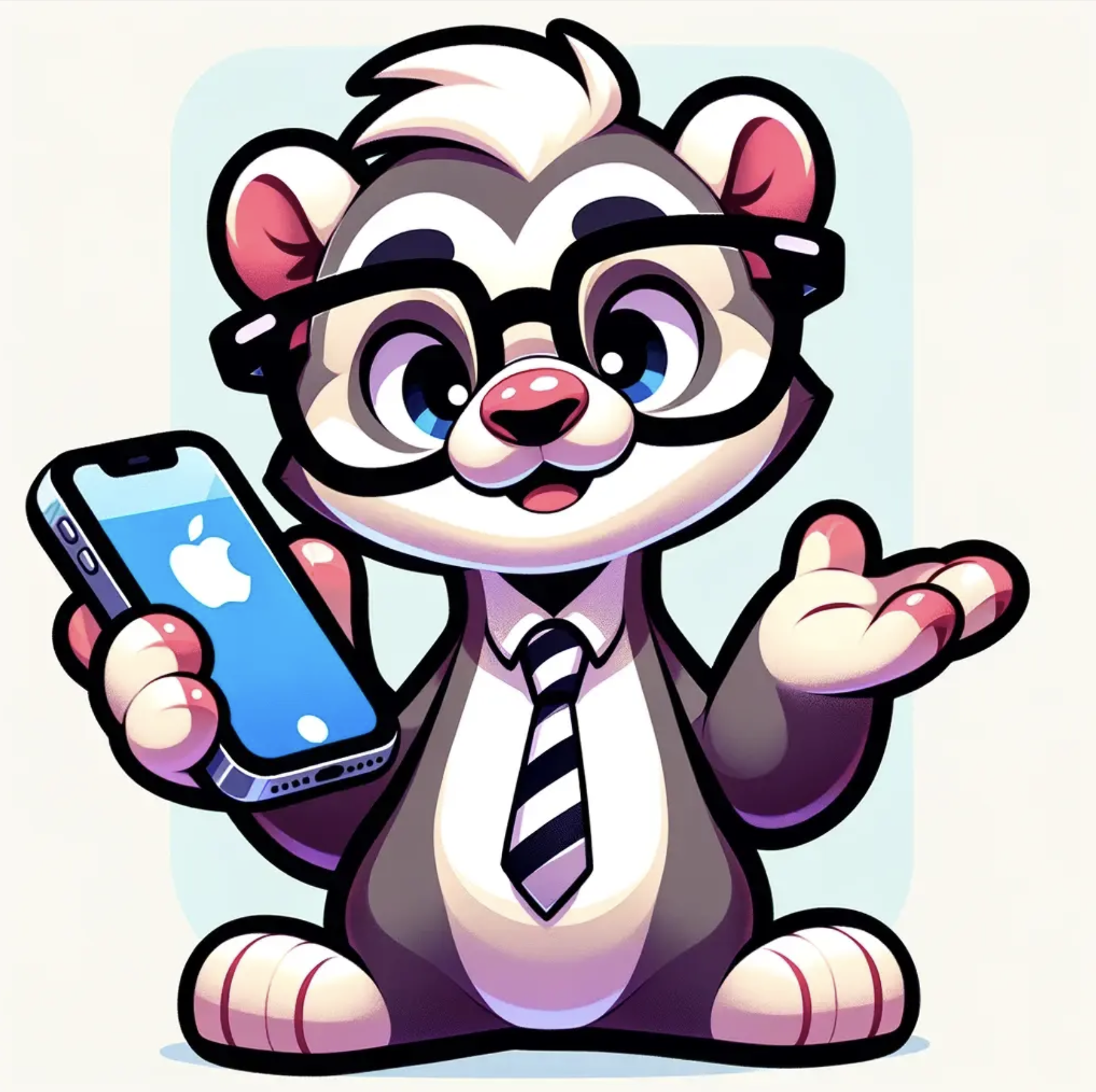} Ferret-UI: Grounded Mobile UI Understanding with Multimodal LLMs} 

\titlerunning{Ferret-UI: Grounded Mobile UI Understanding with Multimodal LLMs}

\author{Keen You, Haotian Zhang, Eldon Schoop, Floris Weers, \\ Amanda Swearngin, Jeffrey Nichols, Yinfei Yang, and Zhe Gan}

\authorrunning{K.~You et al.}

\institute{Apple \\
\email{\{k\_you,haotian\_zhang2,eldon,fweers, \\ aswearngin,jwnichols,yinfeiy,zhe.gan\}@apple.com}}

\maketitle

\begin{abstract}
    Recent advancements in multimodal large language models (MLLMs) have been noteworthy, yet, these general-domain MLLMs often fall short in their ability to comprehend and interact effectively with user interface (UI) screens. In this paper, we present Ferret-UI, a new MLLM tailored for enhanced understanding of mobile UI screens, equipped with \emph{referring}, \emph{grounding}, and \emph{reasoning} capabilities. Given that UI screens typically exhibit a more elongated aspect ratio and contain smaller objects of interest (\emph{e.g.}, icons, texts) than natural images, we incorporate ``any resolution'' on top of Ferret to magnify details and leverage enhanced visual features. Specifically, each screen is divided into 2 sub-images based on the original aspect ratio (\emph{i.e.}, horizontal division for portrait screens and vertical division for landscape screens). Both sub-images are encoded separately before being sent to LLMs. We meticulously gather training samples from an extensive range of elementary UI tasks, such as \emph{icon recognition}, \emph{find text}, and \emph{widget listing}. These samples are formatted for instruction-following with region annotations to facilitate precise referring and grounding. To augment the model's reasoning ability, we further compile a dataset for advanced tasks, including detailed description, perception/interaction conversations, and function inference. After training on the curated datasets, Ferret-UI exhibits outstanding comprehension of UI screens and the capability to execute open-ended instructions. For model evaluation, we establish a comprehensive benchmark encompassing all the aforementioned tasks. Ferret-UI excels not only beyond most open-source UI MLLMs, but also surpasses GPT-4V on all the elementary UI tasks.
\end{abstract}
\begin{figure}[t!]
    \centerline{
    \includegraphics[width=1.1\linewidth]{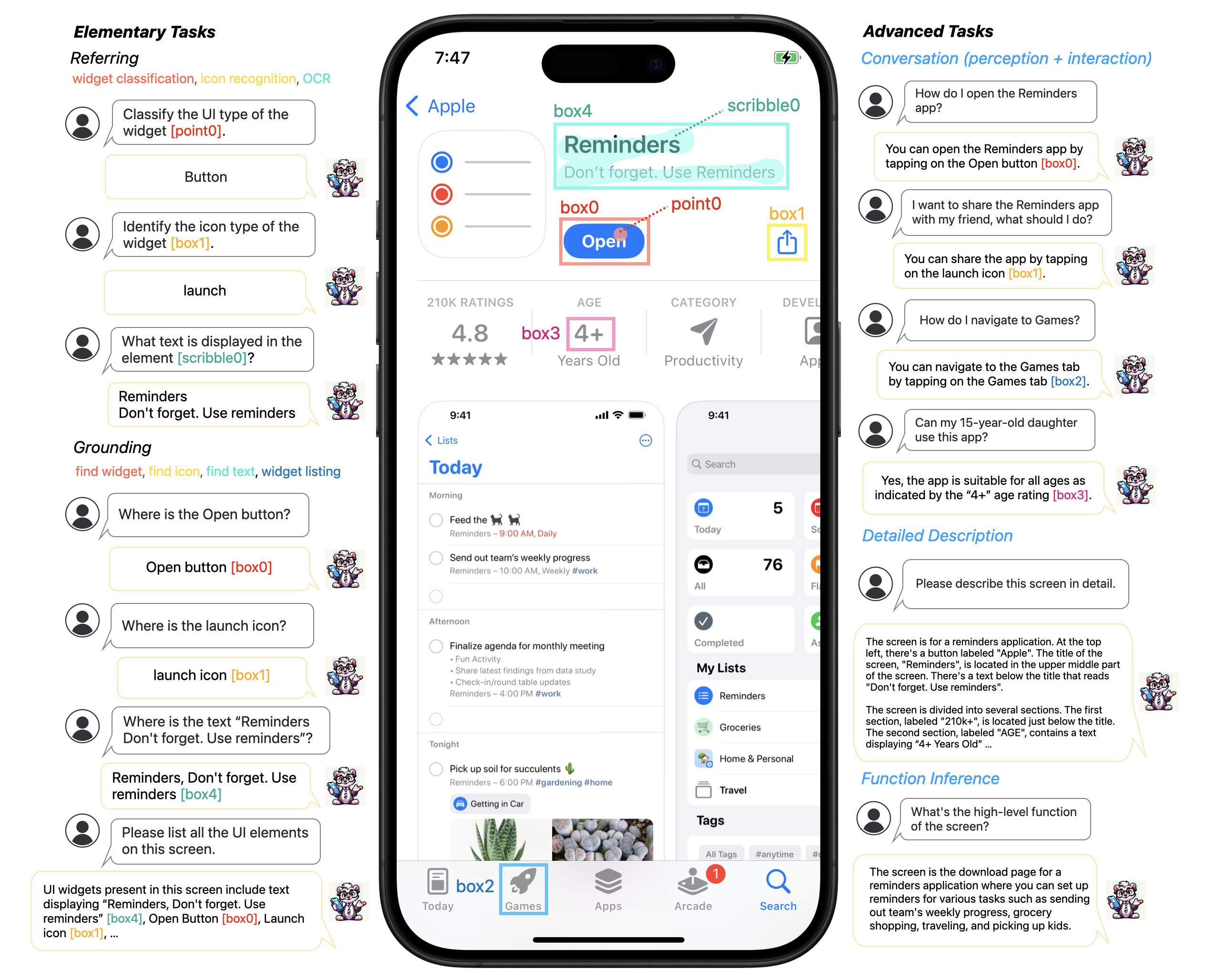}
    }
    \caption{\modelname is able to perform \emph{referring} tasks (\emph{e.g., \textcolor{WildStrawberry}{widget classification}, \textcolor{Goldenrod}{icon recognition}, \textcolor{Aquamarine}{OCR})} with flexible input formats (point, box, scribble) and \emph{grounding} tasks (\emph{e.g., \textcolor{WildStrawberry}{find widget}, \textcolor{Goldenrod}{find icon}, \textcolor{Aquamarine}{find text}, \textcolor{NavyBlue}{widget listing)}} on mobile UI screens. These elementary tasks equip the model with rich visual and spatial knowledge, enabling it to distinguish UI types at both coarse and fine levels, such as between various icons or text elements. This foundational knowledge is crucial for performing more advanced tasks. Specifically, \modelname is able to not only discuss visual elements in \emph{\textcolor{cyan}{detailed description}} and \emph{\textcolor{cyan}{perception conversation}}, but also propose goal-oriented actions in \emph{\textcolor{cyan}{interaction conversation}} and deduce the overall function of the screen via \emph{\textcolor{cyan}{function inference}}.} 
    \label{fig:opening_example}
    \vspace{-3mm}
\end{figure}

\section{Introduction}
Mobile applications have become an important part of daily life, serving as tools for individuals to achieve personal goals including searching for information, making reservations, and seeking entertainment. In this usage, we inspect the current screen visually, and perform the desired actions based on our goals. Automating this process of perception and interaction has the potential to help users achieve their goals with relative ease. Moreover, it is also a valuable building block for accessibility \cite{edwards1995access}, multi-step UI navigation \cite{hong2023cogagent,zhang2023appagent,wang2024mobileagent}, app testing \cite{amalfitano2011gui,linares2017continuous}, usability studies \cite{jiang2018usability}, and many others. 

To facilitate seamless automation of perception and interaction within user interfaces, a sophisticated system endowed with a set of key capabilities is essential. Such a system must possess the ability to not only comprehend the entirety of a screen but also to concentrate on specific UI elements within that screen. With visual understanding as the foundation, it should further be able to map natural language instructions to corresponding actions within a given UI, execute advanced reasoning, and provide exhaustive details concerning the screens it interacts with. These requirements necessitate the development of a vision-language model adept at both referring and grounding in relation to UI screens. Here, \emph{referring} requires the system to utilize particular regional image information in the screen input, while \emph{grounding} involves the model's capacity to identify and denote precise locations on the screen in its outputs.

Existing approaches are insufficient in fully addressing these key capabilities. On one hand, while Multimodal Large Language Models (MLLMs) like Ferret~\cite{you2023ferret}, Shikra~\cite{chen2023shikra}, and Kosmos2~\cite{peng2023kosmos} demonstrate strong referring and grounding capabilities, their scope is mainly restricted to natural images. Directly adapting these models to UI screens can be limiting, since UI screens typically exhibit more elongated aspect ratios and contain smaller objects of interests (\emph{e.g.}, icons and texts) than natural images. Relying solely on a directly resized, low-resolution global image could lead to loss of important visual signals that are essential for screen understanding and interaction. On the other hand, other works targeting directly at UI tasks have primarily focused on processing entire screens as singular inputs (\emph{e.g.}, Pix2Struct~\cite{lee2023pix2struct}, ILuvUI~\cite{jiang2023iluvui}, CogAgent~\cite{hong2023cogagent}), only supports referring tasks with one bounding box in the input (\emph{e.g.}, Spotlight~\cite{li2023spotlight}), and leveraging GPT-4V~\cite{yang2023dawn} to navigate UI screens, as seen in MM-Navigator~\cite{yan2023gpt}, AppAgent~\cite{zhang2023appagent}, and MobileAgent~\cite{wang2024mobileagent}. Furthermore, the tasks studied in these work do not comprehensively cover all dimensions of UI screen understanding. 

\begin{figure}[t!]
    \centerline{
        \includegraphics[width=\textwidth]{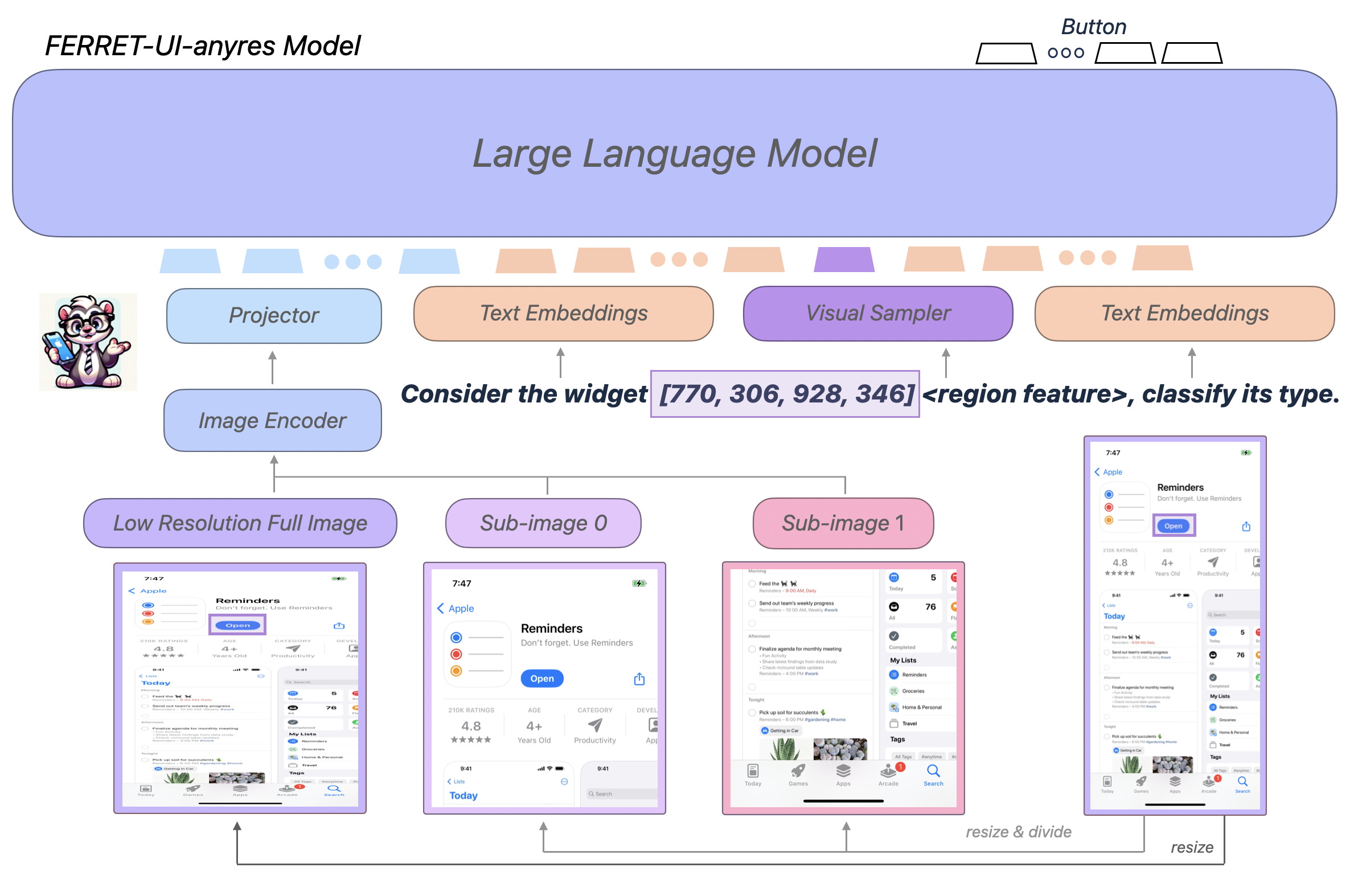}
    }
    \caption{Overview of Ferret-UI-anyres architecture. While Ferret-UI-base closely follows Ferret's architecture, Ferret-UI-anyres incorporates additional fine-grained image features. Particularly, a pre-trained image encoder and projection layer produce image features for the entire screen. For each sub-image obtained based on the original image aspect ratio, additional image features are generated. For text with regional references, a visual sampler generates a corresponding regional continuous feature. The LLM uses the full-image representation, sub-image representations, regional features, and text embeddings to generate a response.}
    \label{fig:ferret-ui-architecture}
    \vspace{-3mm}
\end{figure}

In this paper, we present Ferret-UI, the first MLLM designed to execute precise referring and grounding tasks specific to UI screens, while adeptly interpreting and acting upon open-ended language instructions. We address the aforementioned limitations by focusing on three pivotal dimensions: ($i$) improved model architecture, ($ii$) data curation, and ($iii$) benchmark establishment. For model architecture, we base our approach on Ferret \cite{you2023ferret}, an MLLM known for its strong performances in referring and grounding with natural images. We posit that such capabilities provide a solid foundation in interactive UI-centric tasks. For flexible adaptation of UI screen aspect ratios, we integrate ``any resolution'' (anyres) into Ferret similar to \cite{liu2024llavanext, lin2023sphinx, gao2024sphinxx}, but with pre-defined grid configurations to divide the full image into sub-images so that both portrait and landscape screens can be accommodated. As later shown in Fig. \ref{fig:ferret-ui-architecture}, sub-image features are used in addition to global image features to help magnify details and provide enhanced visual features. 

To train Ferret-UI, we generate data at different granularities, covering basic semantic and spatial tasks for UI primitives to advanced reasoning tasks. We first generate training samples for elementary UI tasks using a template-based approach. This encompasses \emph{referring} tasks such as \emph{widget classification}, \emph{icon recognition}, \emph{OCR}, and \emph{grounding} tasks like \emph{find widget}, \emph{find icon}, \emph{find text}, and \emph{widget listing}. These tasks are instrumental in teaching the model to understand the semantics and spatial positioning of UI elements, enabling the model to make distinctions at both a broad level (among various UI types) and a more detailed level (within specific UI types, such as icons or text). For advanced tasks, we use GPT-4~\cite{openai2024gpt4} to generate data, including \emph{detailed description}, \emph{conversation perception}, \emph{conversation interaction}, and \emph{function inference}. These advanced tasks prepare the model to engage in more nuanced discussions about visual components, formulate action plans with specific goals in mind, and interpret the general purpose of a screen. Fig. \ref{fig:opening_example} illustrates examples of Ferret-UI's proficiency in handling the 11 tasks ranging from basic to advanced.

To assess these capabilities, we develop a comprehensive test benchmark featuring 14 diverse mobile UI tasks in terms of referring and grounding. This includes 3 tasks from Spotlight~\cite{li2023spotlight} (\emph{screen2words}, \emph{widget captions}, and \emph{taperception}), and dual versions of the 11 UI tasks previously described, tailored for both iPhone and Android screens. We conduct comprehensive evaluation of a variety of UI understanding models, including both open-source MLLMs (\emph{e.g.}, CogAgent \cite{hong2023cogagent} and Fuyu \cite{fuyu-8b}) and GPT-4V. We observe that Ferret-UI significantly surpasses the base Ferret model, illustrating the importance of domain-specific model training. Compared to GPT-4V, Ferret-UI demonstrates superior performance in elementary UI tasks. Notably, in the context of advanced tasks, Ferret-UI surpasses both Fuyu and CogAgent.

Our contributions are summarized as follows. ($i$) We propose Ferret-UI with ``any-resolution'' (anyres) integrated to flexibly accommodate various screen aspect ratios. It represents the first UI-centric MLLM that is capable of effectively executing referring, grounding, and reasoning tasks. ($ii$) We define a set of elementary and advanced UI tasks, for which we have meticulously gathered training samples for model training. ($iii$) We develop a comprehensive test benchmark encompassing all the tasks under investigation. Through careful experiments and analysis, we offer insights into the model's capabilities and limitations.
\section{Related Work} \label{sec:related_work} 

Earlier works \cite{shi2017world, liu2018reinforcement, gur2018learning, li2020mapping, burns2022dataset} in the area focus on studying simplified web and mobile screens. With recent advances in both LLMs~\cite{touvron2023llama,openai2024gpt4,gu2023mamba,jiang2023mistral, huang2023language, driess2023palm, anil2023palm} and MLLMs~\cite{liu2023llava, zhu2023minigpt, ye2023mplug, li2023otter, dai2023instructblip,sun2023generative,mckinzie2024mm1,li2023multimodal}, the approaches to many research problems have been transformed, including UI understanding. Several works have explored the use of MLLMs for UI tasks. 
Specifically, ILuvUI~\cite{jiang2023iluvui} and Spotlight \cite{li2023spotlight} concentrate on single-screen UI tasks while exploring various UI tasks by fine-tuning on GPT-generated data and delving into UI tasks such as screen summarization and widget interaction. 

MobileAgent \cite{wang2024mobileagent} and AppAgent \cite{zhang2023appagent} represent a different approach, utilizing MLLMs as agents for UI screen navigation, with MobileAgent employing external detection modules for action generation and AppAgent leveraging overlaid UI element IDs and screen XML files for predefined actions. CogAgent \cite{hong2023cogagent}, built upon CogVLM \cite{wang2023cogvlm}, shifts the focus towards using only screen images for complex UI navigation, eliminating the need for UI-specific modules. Here are some more examples among other works that utilize LLMs \cite{kim2023language, zheng2024synapse, deng2024mind2web, gur2023real} and MLLMs \cite{shaw2024pixels, zhan2023you, yan2023gpt, gao2023assistgui, zheng2024gpt, cheng2024seeclick,baechler2024screenai} in the space. 

In this work, we focus on fine-grained mobile UI understanding with MLLMs. Naturally, our work also aligns with the recent burgeoning literature focused on empowering MLLMs for referring and grounding tasks~\cite{zhang2023gpt4roi,chen2023shikra,peng2023kosmos, lai2023lisa,zhao2023bubogpt,you2023ferret,zhang2023llava}.

\section{Method}
Ferret-UI is built upon Ferret \cite{you2023ferret}, which is a MLLM that excells in spatial referring and grounding within natural images of diverse shapes and levels of detail. It can interpret and interact with regions or objects, whether they are specified as points, boxes, or any free-form shapes. Ferret contains a pre-trained visual encoder (\emph{e.g.}, CLIP-ViT-L/14)~\cite{radford2021learning} and a decoder-only language model (\emph{e.g.}, Vicuna~\cite{zheng2023judging}). Furthermore, Ferret incorporates a unique hybrid representation technique that transforms specified regions into a format suitable for processing by the LLM. At its core, a spatial-aware visual sampler is designed to adeptly manage continuous features of region shapes in different sparsity levels. 

To instill UI expert knowledge into Ferret, we make two extensions to develop Ferret-UI: ($i$) the definition and construction of UI referring and grounding tasks (Section~\ref{sec:dataset}); and ($ii$) model architecture adjustment to better deal with screen data. 
Specifically, Ferret-UI includes a broad range of UI referring tasks (\emph{e.g.}, OCR, icon recognition, widget classification) and grounding tasks (\emph{e.g.}, find text/icon/widget, widget listing) for model training, building up a strong UI understanding foundation for advanced UI interactions. Unlike previous MLLMs that require external detection modules or screen view files, Ferret-UI is self-sufficient, taking raw screen pixels as model input. This approach not only facilitates advanced single-screen interactions, but also paves the way for new applications, such as improving accessibility. Initial explorations of the dataset result in two modeling insights: ($i$) UI screens are predominantly characterized by aspect ratios that are more extended compared to those found in natural images, as evidenced in Tab.~\ref{tab:screen_num_distribution}; ($ii$) the tasks involve many objects of interest (\emph{i.e.}, UI widgets like icons and texts) that are significantly smaller than the objects typically observed in natural images. For example, many questions focus on icons that occupy less than 0.1\% of the entire screen. Thus, relying solely on a single directly resized, low-resolution global image could lead to significant loss of visual details. 

To address this problem, we apply the idea of ``any resolution'' (anyres), as advocated in SPHINX~\cite{lin2023sphinx, gao2024sphinxx}, LLaVA-NeXT~\cite{liu2024llavanext}, and Monkey~\cite{li2023monkey}, to Ferret-UI. Specifically, we opt for two grid configurations, 1x2 and 2x1, which are chosen based on the aspect ratios of the original screens as depicted in Tab.~\ref{tab:screen_num_distribution}. Given a screen, the grid configuration that most closely matches its original aspect ratio is selected. Subsequently, the screen is resized to fit the selected grid configuration and is then partitioned into sub-images. Intuitively, portrait screens are divided horizontally, whereas landscape screens are divided vertically. All sub-images are encoded separately using the same image encoder, and the LLM uses all visual features of varying granularity with both the full image context as well as the enhanced details. The overall architecture of Ferret-UI, including the any-resolution adjustments, is illustrated in Fig.~\ref{fig:ferret-ui-architecture}.

\section{Dataset and Task Formulation}\label{sec:dataset}
In this section, we detail the process of generating datasets for model training and evaluation. Specifically, we describe the UI detection data collection process in Section~\ref{sec: ui_data}, and we outline how we create task-specific data from raw detections in Section~\ref{sec: task_formulation}.

\begin{table}[t!]
    \begin{subtable}[h]{0.49\textwidth}
        \centering
        \begin{tabular}{ccrr} \toprule
        \bf Platform & \bf Resolution & \bf Train & \bf Test \\ \midrule
        Android & 2560$\times$1440 & 26,527 & 3,080 \\ \midrule
        \multirow{4}{*}{iPhone} & 1792$\times$828 & 74,953 & 8,297 \\
        & 828$\times$1792  & 4,225  & 461  \\ 
        & 2436$\times$1125 & 5,420  & 635  \\ 
        & 1125$\times$2436 & 87    & 17   \\ \bottomrule
        \end{tabular}
       \caption{Number of screens by resolution.}
       \label{tab:screen_num_distribution}
    \end{subtable}
    \hfill
    \begin{subtable}[h]{0.49\textwidth}
        \centering
        \begin{tabular}{lrr} \toprule
\bf Task   & \bf iPhone & \bf Android \\ 
\midrule
screen2words          & \quad - &      \quad 78k \\
widget captions       & \quad - &      \quad 109k \\
taperception          & \quad - &      \quad 14k \\ \midrule

elementary tasks      & \quad 40k$\times$7 & \quad 40k$\times$7 \\ \midrule
advanced tasks        &  \quad 10k$\times$4 & \quad 10k$\times$4 \\ 
\bottomrule
\end{tabular}
\caption{Number of samples per training task.}
        \label{tab:task_data_num_distribution}
     \end{subtable}
     \caption{Mobile UI screen and training data statistics.}
     \label{tab:ui_screen_statistics}
     \vspace{-4mm}
\end{table}

\subsection{UI Data Collection} \label{sec: ui_data}

\noindent\textbf{UI Screens.}
To build a model capable of perceiving and interacting with mobile screens, it is crucial to gather a varied collection of such screens. This study examines screens from both iPhone and Android devices.

For Android screens, we use a subset of the RICO dataset \cite{deka2017rico}. Specifically, we consider the tasks in Spotlight \cite{li2023spotlight}, whose data is publicly available, including \textit{screen2words}, \textit{widgetcaptions}, and \textit{taperception}. We aggregate unique images for each split (train and test) among the tasks to form our own data splits. 
In total, there are 26,527 train images and 3,080 test images. 

For iPhone screens, we use the AMP dataset \cite{zhang2021screenrecognition}, which spans a broad spectrum of applications. A subset is randomly selected and divided into training and test splits. The iPhone screens come in various sizes, resulting in a total of 84,685 training images and 9,410 test images. The breakdown of image sizes is summarized in Tab. \ref{tab:screen_num_distribution}.

\vspace{1mm}
\noindent\textbf{UI Screen Elements Annotation.}
After collecting Android and iPhone screens, we further collect fine-grained element annotation from screens using a pre-trained pixel-based UI detection model~\cite{zhang2021screenrecognition}. For each detected UI element, the output includes a UI type (Button, Text, Icon, Picture, \emph{etc.}), the corresponding bounding box, and the text displayed on it, if any, identified by the Apple Vision Framework\footnote{https://developer.apple.com/documentation/vision}. 
We further use heuristics from Screen Recognition~\cite{zhang2021screenrecognition} to group individual detections into larger units, \emph{e.g.}, multiple lines of text are merged into one group, an image is grouped with its caption, \emph{etc}.

\subsection{Task Formulation} \label{sec: task_formulation}

This section describes how we convert the UI screens along with the associated detection data to a format that can be used to train an MLLM. We elaborate three different approaches devised for the construction of the dataset.

\begin{figure}[t!]
    \centerline{
        \includegraphics[width=\textwidth]{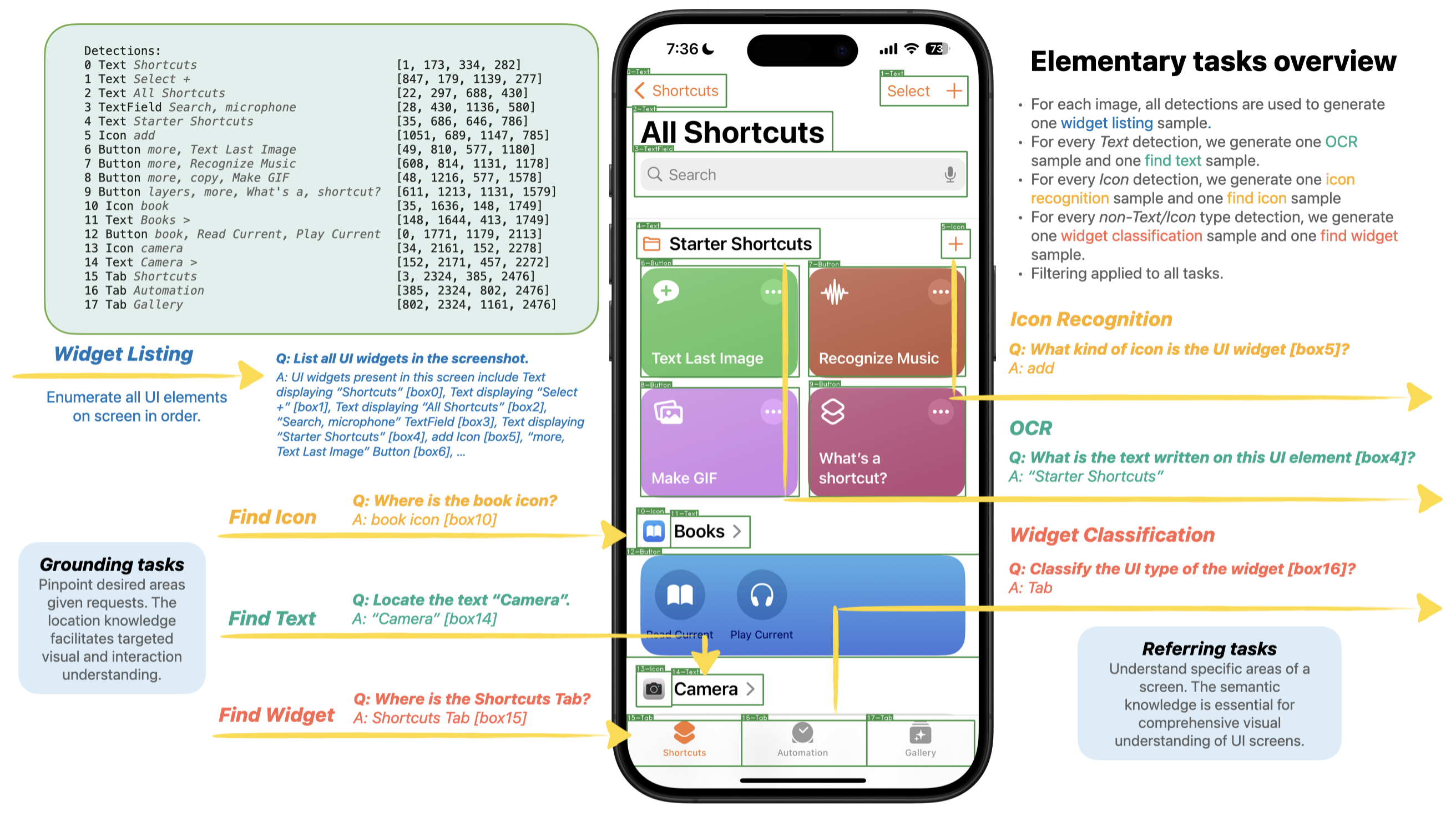}
    }
    \caption{\textbf{Elementary task data generation overview}. A UI detector outputs all detected elements, with each element's \emph{type}, \emph{text}, and \emph{bounding boxes}. These detections are used to create training samples for elementary tasks. For \emph{grounding tasks}, we use all element detections to create one sample for widget listing whereas the remaining tasks focus on one element at a time. We separate the elements into \emph{icons}, \emph{text}, and \emph{non-icon/text widgets}. For each type, we create one referring and one grounding sample.}
    \label{fig:elementary_task_datagen}
\end{figure}

\vspace{1mm}
\noindent\textbf{Reformatting Spotlight.}
We first take \textit{screen2words}, \textit{widgetcaptions}, and \textit{taperception} from the existing Spotlight tasks~\cite{li2023spotlight}, and format them into conversational QA pairs. Specifically, GPT-3.5 Turbo is used to create a varied set of prompts from base prompts we author for respective tasks:
\begin{itemize}
    \item\textbf{Screen2words}\textit{: Provide a summary of this screenshot}; 
    \item\textbf{Widget Captions}\textit{: For the interactive element [bbox], provide a phrase that best describes its functionality};
    \item\textbf{Taperception}\textit{: Predict whether the UI element [bbox] is tappable}.
\end{itemize}
For each training example, we sample a prompt for the corresponding task and pair it with the original source image and ground-truth answer.

\vspace{1mm}
\noindent\textbf{Elementary Tasks.}
In addition to the Spotlight tasks, we use paired screens and UI elements mentioned in Section \ref{sec: ui_data} to generate data for novel UI tasks that rely on grounding and referring capabilities. We introduce 7 tasks using this approach, one set for each of Android and iPhone screens: \emph{OCR}, \emph{icon recognition}, and \emph{widget classification} for \emph{referring}; and \emph{widget listing}, \emph{find text}, \emph{find icon}, and \emph{find widget} for \emph{grounding}.
We define \emph{referring tasks} as the ones with bounding boxes in the inputs, while \emph{grounding tasks} are the ones with bounding boxes in the outputs.

For each task, we also use GPT-3.5 Turbo to expand a base prompt to introduce variants of the task question. Details for data generation are illustrated in Fig. \ref{fig:elementary_task_datagen}. The number of training samples for each task is summarized in Tab. \ref{tab:task_data_num_distribution}. The number of test samples for all tasks are 5K. In experiments, we sample from this pool of training data with different ratios to construct our training data mixture. 

\begin{figure}[t!]
    \centerline{
        \includegraphics[width=0.98\textwidth]{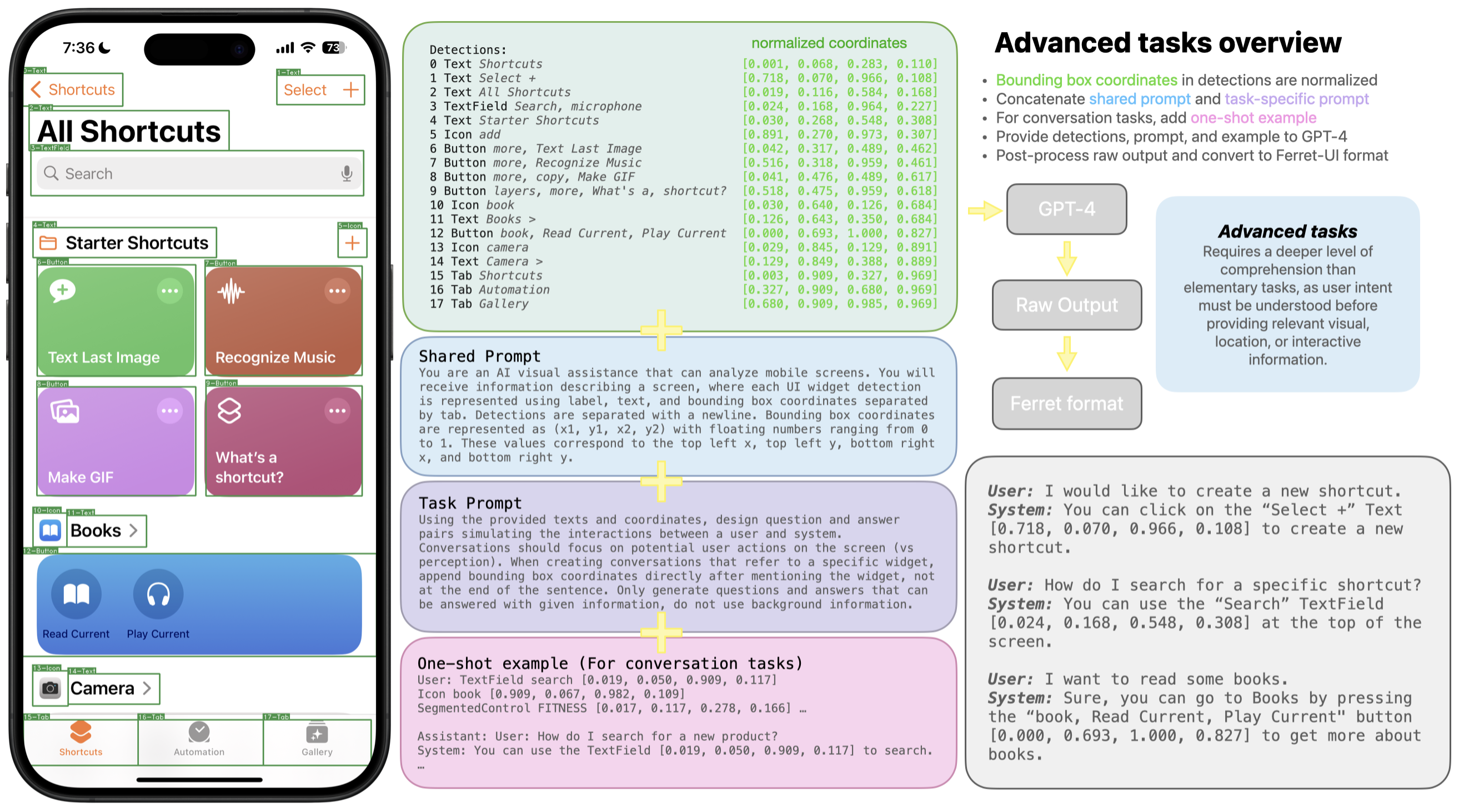}
    }
    \caption{\textbf{Advanced task data generation overview.} We first normalize bounding box coordinates from the detection outputs, then we send the detections, prompts, and optional one-shot example to GPT-4. For detailed description and function inference, we pair the generated response with a pre-selection of prompts to train Ferret-UI. For conversation tasks, we directly transform GPT-4 output to multi-turn conversations.}
    \label{fig:advanced_task_data_gen}
    \vspace{-5mm}
\end{figure}

\vspace{1mm}
\noindent\textbf{Advanced Tasks.}
To incorporate reasoning abilities into our model, we follow LLaVA~\cite{liu2023llava}, and additionally collect data of 4 more formats using GPT-4. We focus on iPhone screens for this part of the data collection, filtering our examples to those with more than 2 but fewer than 15 detections. These examples are sent together with prompts to GPT-4 to create data of the desired format---the actual images are not used. Fig. \ref{fig:advanced_task_data_gen} illustrates the training data generation process for advanced tasks.

The four tasks are \emph{detailed description}, \emph{conversation perception}, \emph{conversation interaction}, and \emph{function inference}. Among these, we expand base prompts for detailed description and function inference to pair them with the GPT-4 response as the input data in our model training. For conversations, we provide an in-context example for GPT-4 to better follow bounding box formats in its output. From the raw GPT-4 output, we parse the bounding boxes and transform them into the correct multi-turn conversation format for our model. In total, we have created 40K valid conversations from GPT-4 generated data. More details about our data collection pipeline and detailed analysis of our collected data are provided in the Appendix.

While our training data collection primarily targets iPhone screens, we assemble test sets for both iPhone and Android platforms. For each task, we select 25 test screens from iPhone and 5 from Android. Due to overlaps in images across different tasks, the total number of unique images amounts to 56 for iPhone and 13 for Android. For evaluation, we randomly select 2 QA pairs for the conversational tasks, creating two distinct test instances with precisely one question in each input. Utilizing these test images, we formulate 20/40/38/20 questions for iPhone and 5/10/10/10 questions for Android, for the four tasks, respectively.
\section{Experiments}

We first present our main results in Section~\ref{sec:main_results}, followed by ablation studies in Section~\ref{sec:ablation_studies}. Then, detailed analysis of results on elementary and advanced UI tasks is provided in Section~\ref{sec:analysis_1} and \ref{sec:analysis_2}, respectively.

\vspace{1mm}
\noindent \textbf{Setup.} In this section, Ferret-UI-anyres refers to the version with any-resolution integrated, Ferret-UI-base refers to the version directly following the Ferret architecture, and Ferret-UI refers to both configurations. During training, both the decoder and the projection layer are updated while the vision encoder is kept frozen. All the training data is formatted into the instruction-following format, and the training objective is the same as in Ferret. In total, our training mixture has 250K samples. Ferret-UI-base takes 1 day to train while Ferret-UI-anyres takes about 3 days on 8 A100 GPUs.

\begin{table}[t!]
\centering
\scriptsize
\begin{tabularx}{\textwidth}{l *{3}{>{\centering\arraybackslash}X} *{4}{>{\centering\arraybackslash}X}  *{2}{>{\centering\arraybackslash}X}}
\toprule
& \multicolumn{3}{c}{Public Benchmark} & \multicolumn{4}{c}{Elementary Tasks} & \multicolumn{2}{c}{Advanced Tasks} \\
\cmidrule(lr){2-4} \cmidrule(lr){5-8} \cmidrule(lr){9-10}
& S2W & WiC & TaP & Ref-i & Ref-A & Grd-i & Grd-A & iPhone & Android \\
\midrule
Spotlight~\cite{li2023spotlight} 
& 106.7 & 141.8 & \textbf{88.4} & - & - & - & - & - & - \\
Ferret~\cite{you2023ferret} 
& 17.6 & 1.2 & 46.2 & 13.3 & 13.9 & 8.6 & 12.9 & 20.0 & 20.7 \\
Ferret-UI-base 
& 113.4 & \textbf{142.0} & 78.4 & 80.5 & \textbf{82.4} & 79.4 & 83.5 & 73.4 & 80.5 \\
Ferret-UI-anyres & \textbf{115.6} & 140.3 & 72.9 & \textbf{82.4} & \textbf{82.4} & \textbf{81.4} & \textbf{83.8} & 93.9 & 71.7 \\
\midrule
GPT-4V~\cite{achiam2023gpt} 
& 34.8 & 23.5 & 47.6 & 61.3 & 37.7 & 70.3 & 4.7 & \textbf{114.3} & \textbf{128.2} \\

\bottomrule
\end{tabularx}
\vspace{1mm}
\caption{Results of Ferret-UI and baseline models. \emph{S2W}: screen2words, \emph{WiC}: widget captions, \emph{TaP}: taperception. We report the CIDEr score for S2W and WiC and F1 for TaP. For elementary and advanced tasks, we report the averaged performance of corresponding tasks. ``i'': iPhone, ``A'': Android, ``Ref'': Referring, ``Grd'': Grounding.}
\label{tab:main_results}
\vspace{-4mm}
\end{table}

\subsection{Results} \label{sec:main_results}

\begin{figure}[t!]
    \centering
    \includegraphics[scale=0.3]{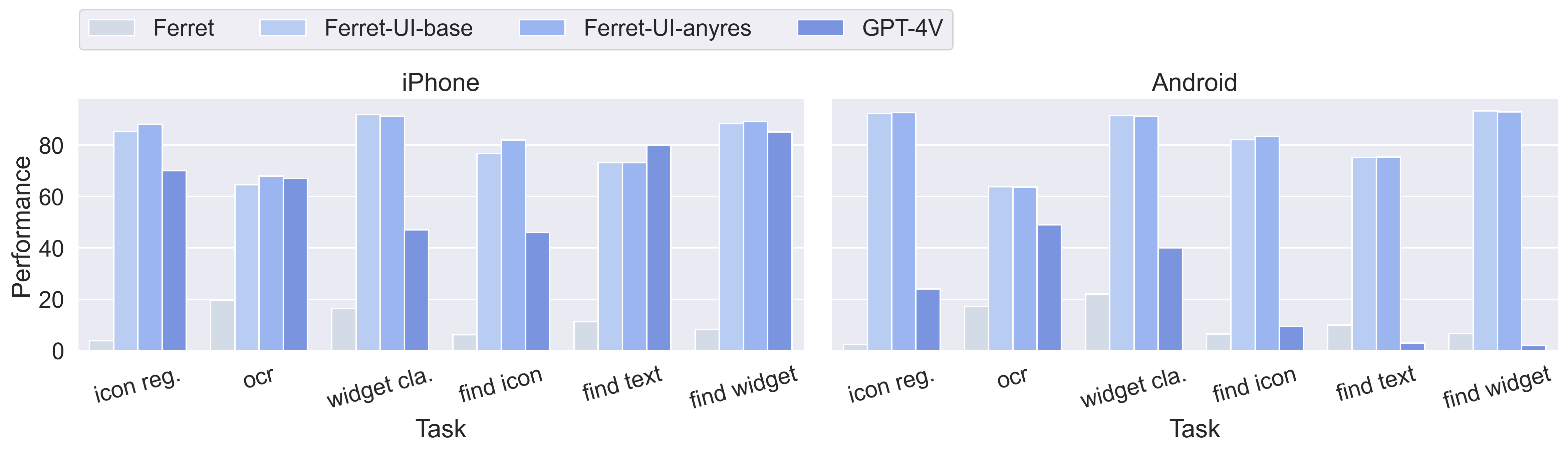}
    \caption{Elementary task performance comparison. Numerous small widgets present on the Android screen make it more challenging for referring and grounding, while Ferret-UI continues to outperform Ferret and GPT-4V on almost all the elementary tasks.}
    \label{fig:elementary_task_perf}
\end{figure}

We compare the performances of Ferret-UI-base, Ferret-UI-anyres, Ferret\footnote{For Ferret, we include the pre-defined classes for icon classification and widget classification in the prompts while the remaining prompts are the same as Ferret-UI.}, and GPT-4V for all tasks. We also include Fuyu~\cite{fuyu-8b} and CogAgent's~\cite{hong2023cogagent} performance on advanced tasks.\footnote{For GPT-4V, we sample a random subset of 100 instances for the Spotlight and elementary tasks for cost efficiency. For GPT-4V evaluation, we follow~\cite{yang2023set} by overlaying indexed bounding boxes of UI elements as visual prompts. Consequently, in grounding tasks, GPT-4V is enabled to make selections from among these candidate boxes. We detail the effort in the Appendix.} Results are summarized in Tab. \ref{tab:main_results}, where the average performance within a category is reported. Performance breakdown for elementary and advanced tasks is shown in Fig. \ref{fig:elementary_task_perf} and Tab. \ref{Tab:advanced_task_perf}, respectively.

\begin{table}[t!]
    \centering
    \scriptsize
    \begin{tabularx}{\textwidth}{l *{5}{>{\centering\arraybackslash}X}  *{5}{>{\centering\arraybackslash}X}}
    \toprule
    & \multicolumn{5}{c}{iPhone} & \multicolumn{5}{c}{Android} \\
    \cmidrule(lr){2-6} \cmidrule(lr){7-11}
    & DetDes & ConvP & ConvI & FuncIn & \textbf{Avg} & DetDes & ConvP & ConvI & FuncIn & \textbf{Avg}\\
    \midrule
    Ferret~\cite{you2023ferret} & 2.5 & 34.7 & 23.7 & 19.1 & 20.0 & 2.0 & 33.9 & 24.9 & 21.9 & 20.7 \\
    Fuyu~\cite{fuyu-8b} & 5.0 & 24.6 & 18.8 & 35.7 & 21.0 & 2.0 & 20.8 & 44.5 & 36.1 & 25.9 \\
    CogAgent~\cite{hong2023cogagent} & 53.1 & 59.7 & 74.8 & 71.9 & 64.9 & 28.0 & 58.5 & 90.1 & \textbf{90.5} & 66.8 \\
    Ferret-UI-base & 64.5 & 75.0 & 77.5 & 76.5 & 73.4 & 90.8 & 72.8 & 79.3 & 79.2 & 80.5 \\
    Ferret-UI-anyres & \textbf{97.4} & 92.1 & 91.1 & \textbf{95.2} & 93.9 & 86.4 & 70.3 & 50.2 & 77.3 & 70.1 \\
    \midrule
    GPT-4V~\cite{achiam2023gpt} & 66.8 & \textbf{105.6} & \textbf{198.5} & 86.3 & \textbf{114.3} & \textbf{126.6} & \textbf{109.4} & \textbf{188.6} & 88.3 & \textbf{128.2} \\
    \bottomrule
    \end{tabularx}
    \vspace{1mm}
    \caption{Advanced task performance comparison. \emph{DetDes}: detailed description, \emph{ConvP}: conversation perception, \emph{ConvI}: conversation interaction, \emph{FuncIn}: function inference.
    }
    \label{Tab:advanced_task_perf}
    \vspace{-3mm}
\end{table}

\vspace{1mm}
\noindent \textbf{Public Benchmark from Spotlight~\cite{li2023spotlight}.}
Compared to Spotlight, Ferret-UI demonstrates superior performance in \emph{S2W} and \emph{WiC}, even though Spotlight uses 80M web page screenshots and 2.69M mobile screenshots for pre-training. Ferret-UI performance falls short on \emph{TaP} but is still competitive; our studies further suggest that this could be due to the noisiness of the taperception labels. Detailed analysis is provided in the Appendix.

\vspace{1mm}
\noindent \textbf{Results on Elementary UI Tasks.}
The average performance of all referring and grounding tasks is summarized in Tab. \ref{tab:main_results}, and the performance breakdown for each task is shown in Fig. \ref{fig:elementary_task_perf}. For referring tasks, we report exact match accuracy for OCR and accuracy for icon recognition and widget classification. For each grounding task, we also report the accuracy, where a correct bounding box is one that has an Intersection-over-Union (IoU) with the label greater than the threshold (0.5). Widget listing performance is not included in the average as we treat it as an auxiliary task.

Ferret-UI outperforms Ferret and GPT-4V in most elementary tasks except for iPhone \emph{find text}. While GPT-4V demonstrates decent performance on iPhone tasks, its performances on Android tasks, especially grounding tasks, are significantly worse. Examining the predictions shows that Android screens have more numerous and smaller widgets, making the grounding tasks more challenging. Furthermore, Ferret-UI's zero-shot performance on the Referring Expression Comprehension task from UIBert~\cite{bai2021uibert} is 76\% when we frame it as the \emph{find widget} task. Notably, with anyres added to Ferret-UI-base, iPhone referring and grounding tasks improve by 2 points.

\vspace{1mm}
\noindent \textbf{Results on Advanced Tasks.} 
The breakdown of task performance for advanced tasks is shown in Tab. \ref{Tab:advanced_task_perf}. As the advanced tasks require open-ended responses, we use GPT-4 to score both the label and the prediction. We report \textit{score for prediction} over \textit{score for label} as a percentage. 

Ferret-UI exhibits commendable performance on advanced tasks for both platforms, despite the absence of Android-specific data in its training dataset. This suggests a notable transferability of UI knowledge across different operating systems. While Fuyu~\cite{fuyu-8b} tends to generate answers that are generally relevant, its responses lack the detail and precision exhibited by Ferret-UI. Conversely, GPT-4V secures higher scores across all tasks by consistently delivering more detailed responses than Ferret-UI, a characteristic that aligns with the preferences of the model evaluator (GPT-4). With Ferret-UI-anyres, iPhone advanced tasks see a huge performance boost of 20 points while Android advanced tasks see a performance drop. As Android advanced task data is not included in the training mix, it could be that as the model gains enriched knowledge about iPhone screen understanding, it loses a bit of generalizability. 

\subsection{Ablation Studies} \label{sec:ablation_studies}
\noindent \textbf{Ablation on Advanced Tasks.}
The design motivation behind elementary tasks is to enhance the model's visual and spatial understanding of basic UI elements. We propose that this enhanced understanding can aid in performing more complex tasks. This hypothesis is examined by investigating how elementary tasks influence the model's ability to handle advanced tasks, with findings detailed in Tab. \ref{advanced_task_ablation}. We see that with only advanced task data, the performance is 64\% for both platforms. The performance of advanced tasks on iPhone shows a consistent improvement of 5\% with the addition of either iPhone or Android elementary tasks. Similarly, adding elementary tasks from the iPhone enhances Android's performance on advanced tasks by about 4\%, whereas incorporating Android elementary tasks boosts this performance by 9\%. Including both iPhone and Android elementary tasks further improves performance by 3\% and 5\% for iPhone and Android advanced tasks, respectively, beyond the improvements seen with a single set of elementary tasks. These observations support our hypothesis that elementary tasks provide the model with enhanced visual and spatial understanding that facilitates advanced tasks.

\begin{table} [!t]
    \begin{subtable}[h]{0.45\textwidth}
        \centering
        \scriptsize
        \begin{tabular}{lcc} \toprule
            & \textbf{iPhone} & \textbf{Android}  \\
        \midrule
        Adv. task only  & 64.6  & 64.3    \\
        + iPhone elem. & 70.3  & 68.6 \\
        + Android elem. & 70.2  & 75.3 \\ 
        + both as in \ref{tab:main_results} & \textbf{73.4}  & \textbf{80.5} \\ \bottomrule
        \end{tabular}
        \caption{\textbf{Advanced Tasks Ablation.} Performance on iPhone and Android advanced tasks. The training configurations are mixing advanced tasks with no other data, with iPhone elementary tasks only, Android elementary tasks only, or both.}
        \label{advanced_task_ablation}
    \end{subtable}
    \hspace{5mm}
    \begin{subtable}[h]{0.45\textwidth}
        \centering
        \scriptsize
        \begin{tabular}{lccc} \toprule
             & S2W & WiC & TaP\\ \midrule
            Spotlight~\cite{li2023spotlight} & 106.7 & 141.8 & \textbf{88.4}  \\ \midrule
            Balanced TaP labels & 111.7 & 133.8 & 76.5 \\
            Spotlight tasks only & 111.3 & 138.7 & 77.6\\
            + Android elem. tasks & 111.3 & 138.0 & 76.8 \\
            + iPhone elem. tasks & 112.4 & 138.9 & 74.8 \\
            + both & 111.3 & 138.7 & 76.0 \\ 
            Full mixture from \ref{tab:main_results} & \textbf{113.4} & \textbf{142.0} & 78.4 \\ \bottomrule
            \end{tabular}
            \caption{\textbf{Spotlight Tasks Ablation.} Performance on \emph{S2W}, \emph{WiC}, \emph{TaP} tasks. For the balanced TaP labels experiment, we up-sample the minority class.}
            \label{tab:spotlight_tasks_ablation}
    \end{subtable}
    \caption{Ablation studies on the factors that impact performance on (a) Advanced tasks and (b) Spotlight tasks.}
    \label{tab:ablation_studies}
    \vspace{-4mm}
\end{table}

\vspace{1mm}
\noindent \textbf{Ablation on Spotlight Tasks.}
Motivated by a desire to explore the impact of different data configurations on Spotlight task performance, we specifically investigate whether adding elementary task data could enhance the model performance, given that these tasks are designed to improve the visual and spatial comprehension of screens. As shown in Tab. \ref{tab:spotlight_tasks_ablation}, the addition of elementary task data—whether exclusively from Android, iPhone, or a combination of both—does not significantly alter performance across the three Spotlight tasks. 
This may be attributed to the short and highly specialized UI-centric vocabulary used in responses in elementary tasks, contrasting with the response style demanded by Spotlight tasks.
Optimal results for Spotlight tasks were observed when data from advanced tasks were integrated alongside all elementary tasks, even though the advanced task data was exclusively derived from iPhone screens. Notably, this yields a 4-point boost in CIDEr score  for the widget captions with the inclusion of advanced task data. We postulate that the free-response format of advanced task answers, which necessitates a more sophisticated set of skills for execution, aligns more closely with the requirements of Spotlight tasks. These tasks demand a comprehensive understanding beyond that of recognizing individual UI elements, as is common in elementary tasks. Moreover, executing advanced tasks requires more sophisticated skills than understanding one specific UI element on the screen as in elementary tasks. Thus, it stands to reason that the skill set honed through advanced tasks would be advantageous for tackling Spotlight tasks, which occupy a middle ground in complexity between elementary and advanced tasks. In one word, the structure of the task assumes greater importance than the source platform of the data incorporated.

\begin{figure}[t!]
    \centerline{
        \includegraphics[width=0.9\linewidth]{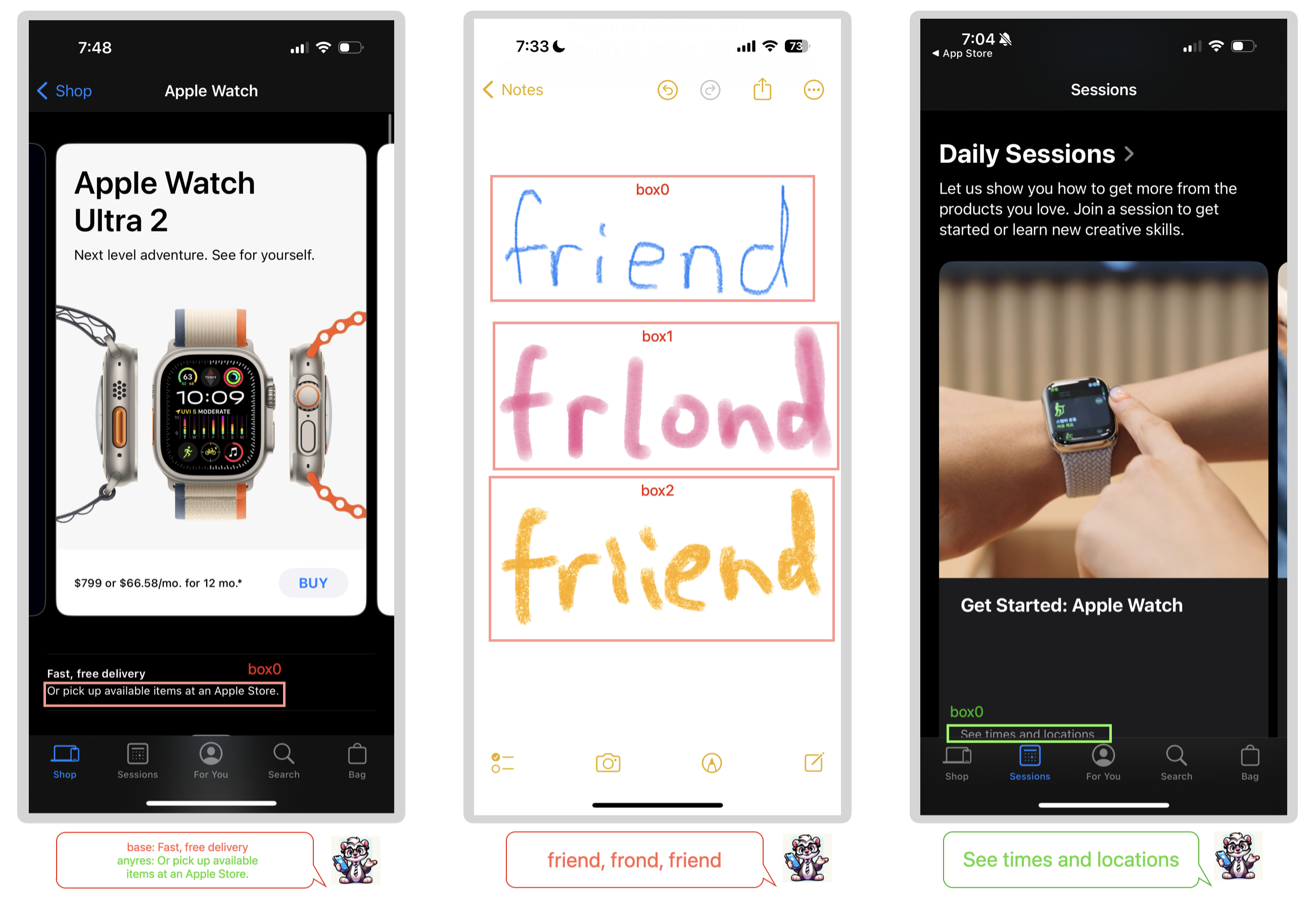}
    }
    \caption{\textbf{OCR Analysis.} \emph{Left}: predict nearby text instead of a targeted region in the base model, corrected in anyres. \emph{Middle}: a tendency to predict valid words. \emph{Right}: Ferret-UI correctly reads cut-off text, while the detection model produces wrong labels.} 
    \label{fig:analyses_ocr}
\end{figure}

\subsection{Result Analysis: Elementary UI Tasks} \label{sec:analysis_1}
\noindent \textbf{Referring Tasks.}
In analyzing Ferret-UI's referring capabilities, we specifically focus on OCR and widget classification predictions. The OCR analysis reveals three notable observations, as depicted in Fig. \ref{fig:analyses_ocr}. First, the model predicts a neighboring text instead of the text in the targeted region. This is common for smaller texts and texts very close to other texts. Remarkably, with anyres integrated, such cases are alleviated, indicating that inputting enlarged sub-images helps the model with smaller visual details.
Second, the model exhibits a tendency to predict actual words rather than merely deciphering characters displayed on the screen. This observation is in line with the semantic-reliance observation of LLMs made in some existing work~\cite{liu2024LMMOCR}. On UI screens, phonetically crafted words that are commonly used as brand titles largely fall under this category. Third, Ferret-UI demonstrates the ability to accurately predict text that is partially cut-off, even in instances where the OCR model returns incorrect texts.

\begin{figure}[t!]
    \centerline{
        \includegraphics[width=0.9\linewidth]{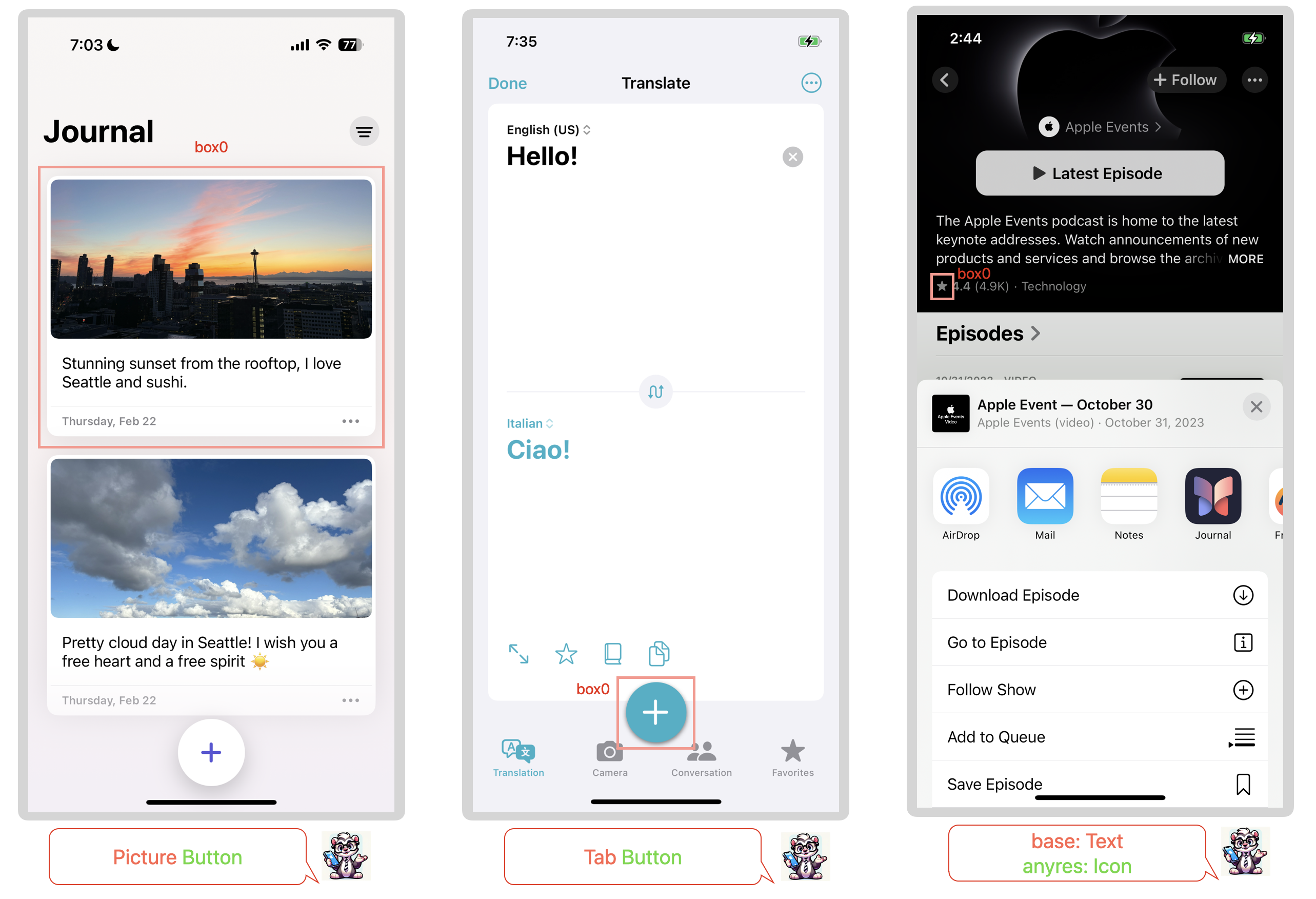}
    }
    \caption{\textbf{Widget Classification Analysis.} \emph{Left}: a large Button consists of a Picture, Icon, and Text misclassified as a Picture. \emph{Middle}: a button seated on top of a row of Tabs misclassified as a Tab. \emph{Right}: a small, text-surrounded icon being classified as text in the base model, but correctly classified with anyres.} 
    \label{fig:analyses_widget_classification}
\end{figure}

Similar to OCR analysis, we show three interesting observations in Fig. \ref{fig:analyses_widget_classification}. First, the model struggles when it needs to understand relationships among widgets. For example, if a large button is made up of a few sub-elements, including Picture, Icon, and text, the model cannot see it as a unified widget but tends to predict it as the sub-element that occupies the largest space. In line with the first observation, when a Tab or an Icon is seated on top of a row of tabs, it is highly likely to be considered part of the tabs. Finally, we discover a common case where small icons surrounded by texts are likely to be predicted as Text, and this is consistent with the observation that small texts tend to be predicted as neighboring texts. With anyres added, such cases are more likely to be predicted correctly, in line with the observation made in OCR.

\begin{figure}[t!]
    \centerline{
        \includegraphics[width=0.9\linewidth]{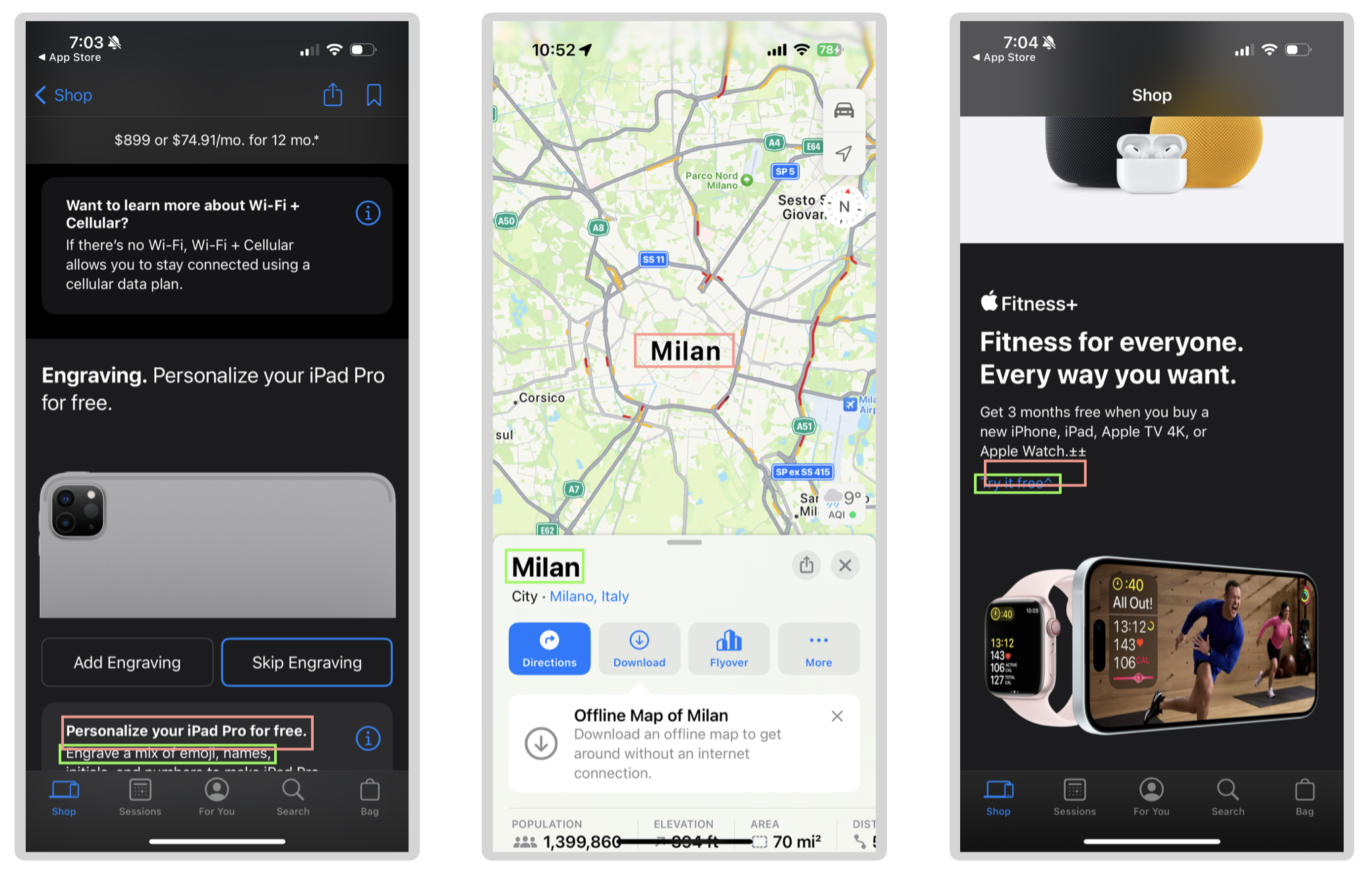}
    }
    \caption{\textbf{Find Text Analysis.} \emph{Left}: a neighboring text is mis-identified as the target. \emph{Middle}: multiple occurrences of the same text. \emph{Right}: predicted boxes not precise.} 
    \label{fig:analyses_find_text}
\vspace{-4mm}
\end{figure}

\vspace{1mm}
\noindent \textbf{Grounding Tasks.}
Using \emph{find text} predictions, as depicted in Fig. \ref{fig:analyses_find_text}, we further elucidate observations from grounding tasks. Echoing the initial observation from the \emph{OCR} analysis, the model may erroneously highlight a piece of text adjacent to the targeted area. Additionally, the occurrence of multiple instances of identical texts suggests the potential for expanding future methods to encompass a range of answers from a singular box to multiple boxes, thereby enhancing the model's utility and accuracy in complex text-finding scenarios.

\begin{figure}[t!]
    \centerline{
    \includegraphics[width=\textwidth]{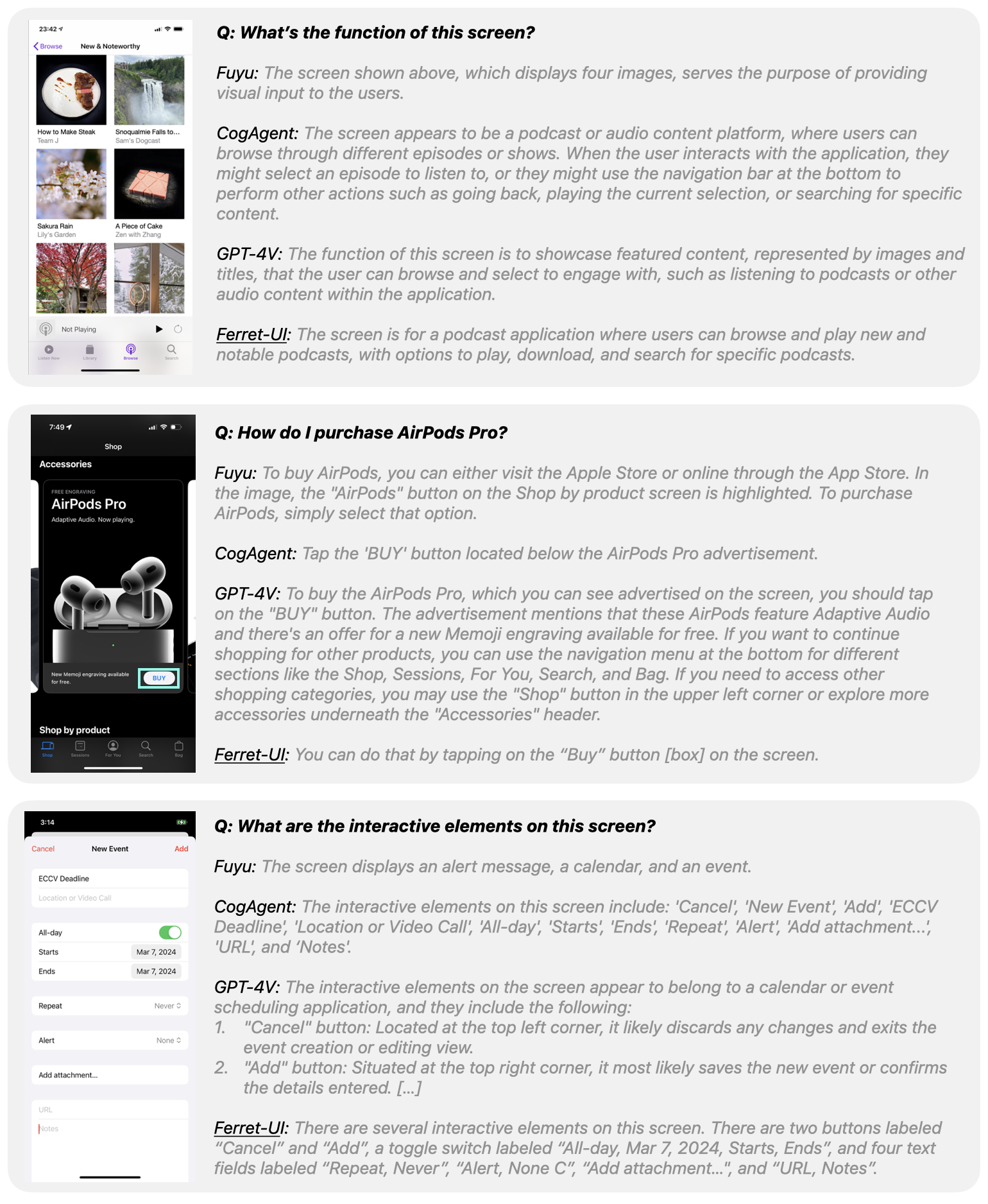}
    }
    \caption{Visualization results of advanced tasks (top to bottom: \emph{function inference}, \emph{conversation interaction}, \emph{conversation perception}) to illustrate the differences among various models (Fuyu vs. CogAgent vs. GPT-4V vs. Ferret-UI).} 
    \label{fig:advanced_task_output}
\end{figure}

\vspace{-1mm}
\subsection{Result Analysis: Advanced UI Tasks} \label{sec:analysis_2}
\noindent \textbf{Grounded Conversation.}
Engaging in grounded conversation is Ferret's unique capability. To better understand the quality of the output bounding boxes in terms of correctness and relevance, we manually grade all output boxes in both Ferret-UI and GPT-4V's \emph{converation interaction} outputs. The accuracies for Ferret-UI and GPT-4V are 91.7\% and 93.4\%, respectively. Considering Ferret-UI generates raw coordinates whereas GPT-4V chooses from a set of pre-defined boxes, Ferret-UI's grounding ability on UI screens is noteworthy. Even though Ferret-UI's received evaluation score falls short of GPT-4V, from inspecting the predictions as in Fig. \ref{fig:advanced_task_output}, we notice that GPT-4V tends to provide extra information that may not be relevant to the question. However, these detailed answers are more favored in scoring than Ferret-UI's concise answers.

\vspace{1mm}
\noindent \textbf{UI detection model is a bottleneck.} 
Given that both our elementary and advanced tasks are predicated upon the detection of UI elements, Ferret-UI is not able to learn aspects of screens that are not detected, such as colors, design, usability, and UI elements that the detection model misses (\emph{e.g.}, topmost time, WIFI, battery). For example, in generating detailed descriptions, GPT-4V is capable of noting ``The overall design conforms to Apple's aesthetic with a minimalistic, clean, dark theme'', a level of insight Ferret-UI is not trained to offer due to its reliance on detected elements alone.

\vspace{1mm}
\noindent \textbf{Set-of-Mark (SoM) Prompting of GPT-4V.}
In our analysis of GPT-4V, the Set-of-Mark (SoM) prompting approach~\cite{yang2023set} is employed, revealing several limitations. First, its effectiveness diminishes in scenarios involving a multitude of small UI elements, a common occurrence in Android detection tasks. The small size of some UI components means that the addition of labels may obscure original content or even extend beyond the intended areas. Second, limiting the assessment to a specified collection of candidate regions restricts the model's ability to reference any given region freely. In the middle example shown in Fig. \ref{fig:advanced_task_output}, the UI detection model treats the entire middle section as one element, covering the texts, image, and the Buy button. Therefore, the model is not able to refer to the ``BUY'' button on its own in its responses, since it is considered part of a collective detection group.
\section{Conclusion}
In this paper, we introduce Ferret-UI, a specialized MLLM designed to enhance comprehension and interaction with mobile UI screens. Through careful design of ``anyres'' to accommodate various screen aspect ratios and curation of training samples that encompass a diverse range of basic and advanced UI tasks, Ferret-UI demonstrates remarkable proficiency in referring, grounding, and reasoning. The advent of these enhanced capabilities promises substantial advancements for a multitude of downstream UI applications, thereby amplifying the potential benefits afforded by Ferret-UI in this domain.

%
%
\bibliographystyle{splncs04}
\bibliography{main}

\newpage
\appendix

\section{Elementary Task Data Generation Details} \label{datagen_details}
Additional details in elementary task data generation are as follows:
\begin{itemize}
  \item In our data generation process, we merge the two distinct classes—``Checked'' and ``Unchecked''—found in the original detection labels for both \emph{Checkboxes} and \emph{Toggles}. 
  \item 
  For widget listing, the answer starts with a common phrase: \textit{UI widgets present in this screen include}. Each element is formatted as ``\{displayed text\} \{UI type\}'' (\emph{e.g.}, ``login button''), except for text elements, which are formatted as ``Text displaying \{displayed text\}''.
  \item For OCR, we consider text with fewer than 10 tokens. If the text is exactly one token, the length needs be to 2 or greater to be included.
  \item For tasks such as \emph{find text}, \emph{find icons}, and \emph{find widget}, it is common to encounter screens containing multiple instances of the same UI element (e.g., multiple login buttons). We employ a filtering mechanism that excludes samples involving UI elements with multiple occurrences within a single screen.
  \item The size of the test set is determined by selecting the smaller value between 5k and the total number of generated test instances.
\end{itemize}

\section{Advanced Task Data Quality Analysis} \label{appendix:conv_analyses}
We conduct a thorough analysis of the quality of our collected data for advanced tasks and provide comprehensive statistics. The vocabulary size for each task is as follows: 30,866 for \emph{detailed description}, 15,666 for \emph{conversation perception}, 12,092 for \emph{conversation interaction}, and 14,896 for \emph{function inference}.

In the realm of \emph{conversation interaction}, we observe 33,649 question turns and 29,933 answer turns. Among these, 15 question turns include bounding boxes, whereas all answer turns include bounding boxes. We compile the most frequently occurring tri-grams for questions and answers in both conversation tasks. Notably, in \emph{conversation perception} questions, the top tri-grams include phrases like \emph{are there any''}, \emph{where is the''}, and \emph{what is the''}, while those for interactions comprise phrases like \emph{How can I''}, \emph{I want to''}, and \emph{Can I do''}. Similarly, in perception answers, prominent tri-grams consist of expressions such as \emph{``bottom of the''}, \emph{``at the top''}, and \emph{``there is a''}, while interaction answers primarily feature tri-grams like \emph{``by tapping on''}, \emph{``tapping on the''}, and \emph{``can tap on''}.

\begin{figure}[h!]
 \begin{subfigure}{0.49\textwidth}
     \includegraphics[width=\textwidth]{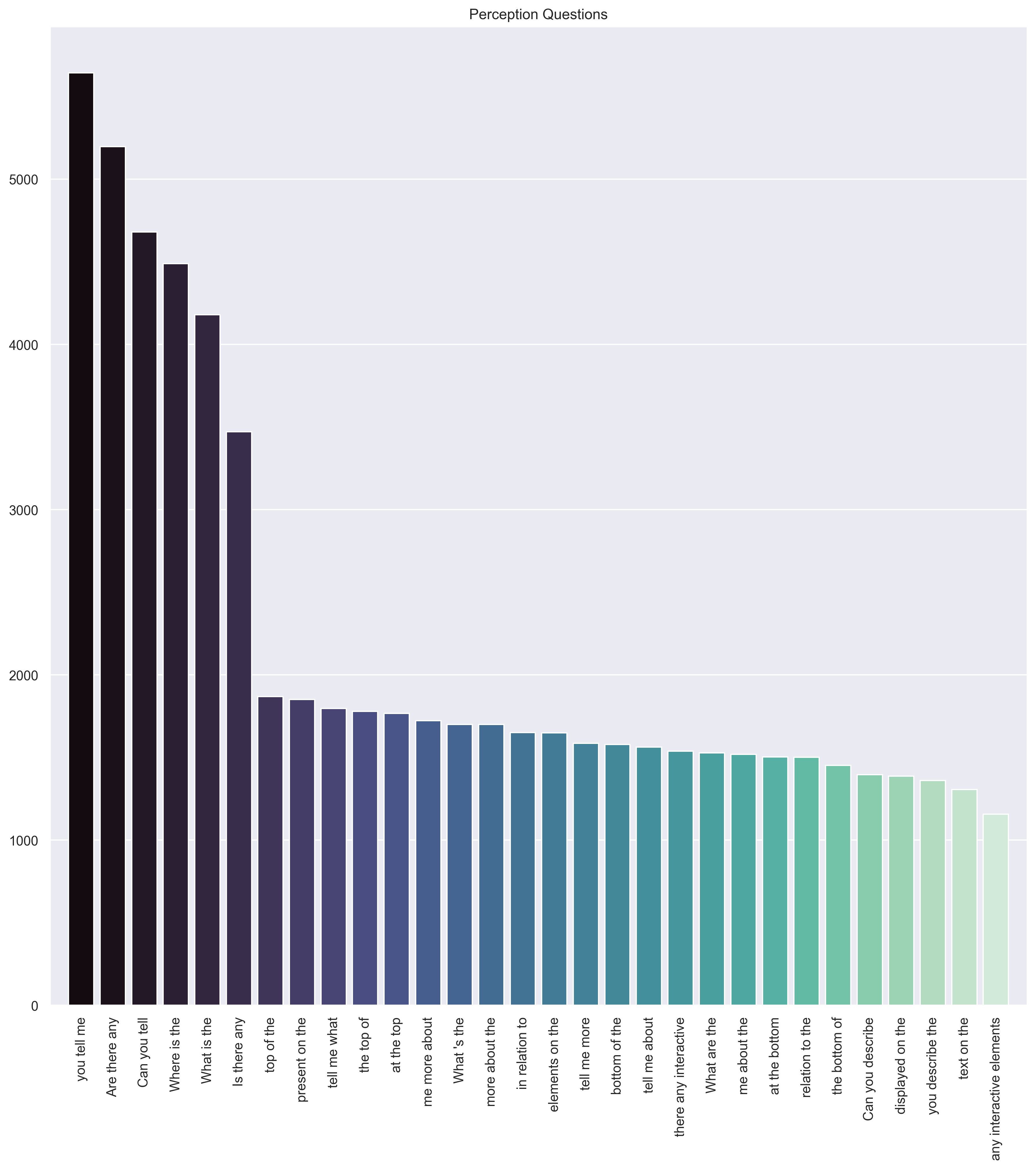}
     \caption{\emph{Conversation perception} \textbf{questions} trigrams distribution.}
     \label{fig:perception_q}
 \end{subfigure}
 \hfill
 \begin{subfigure}{0.49\textwidth}
     \includegraphics[width=\textwidth]{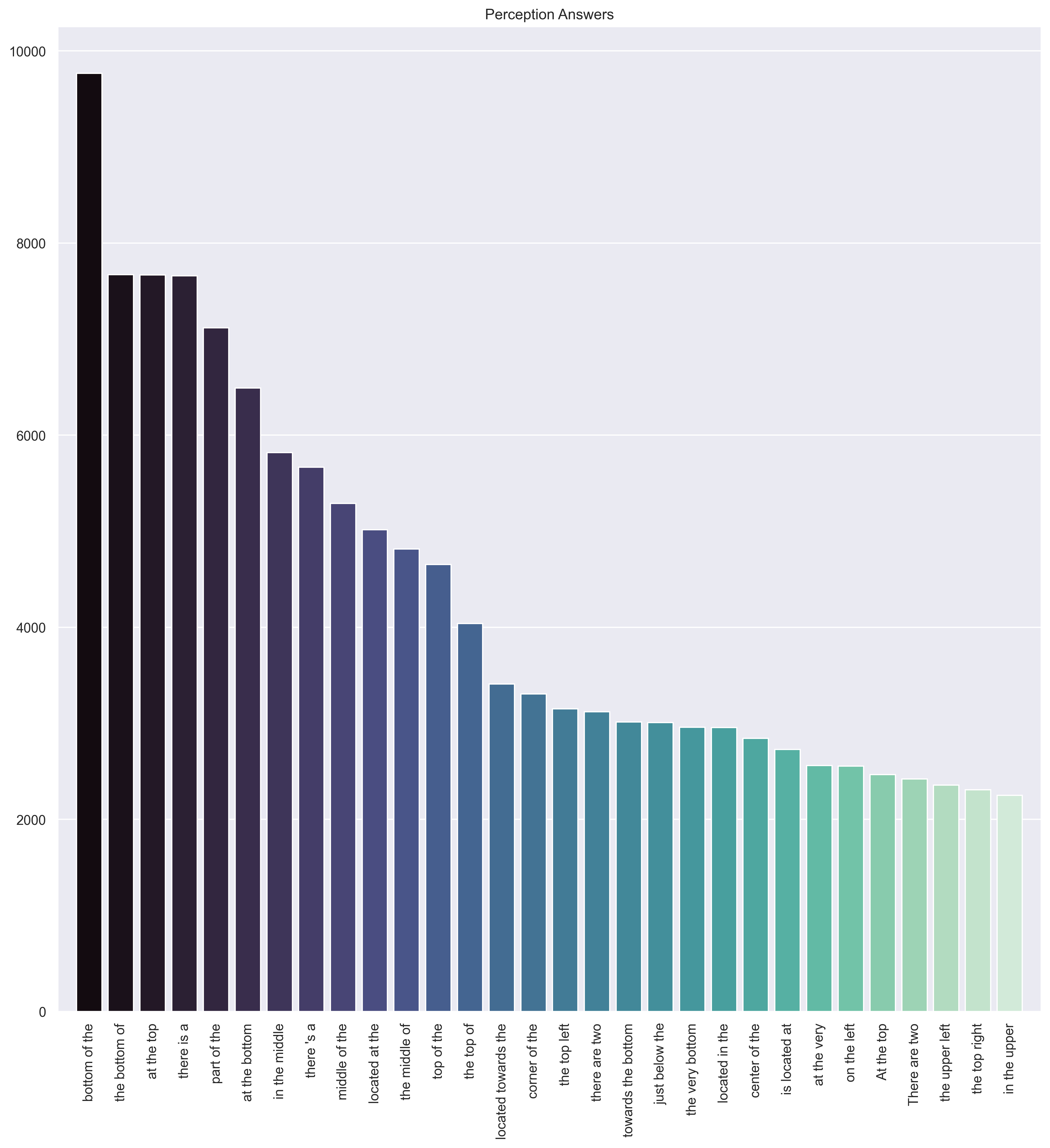}
     \caption{\emph{Conversation perception} \textbf{answers} trigrams distribution.}
     \label{fig:perception_a}
 \end{subfigure}
 
 \medskip
 \begin{subfigure}{0.49\textwidth}
     \includegraphics[width=\textwidth]{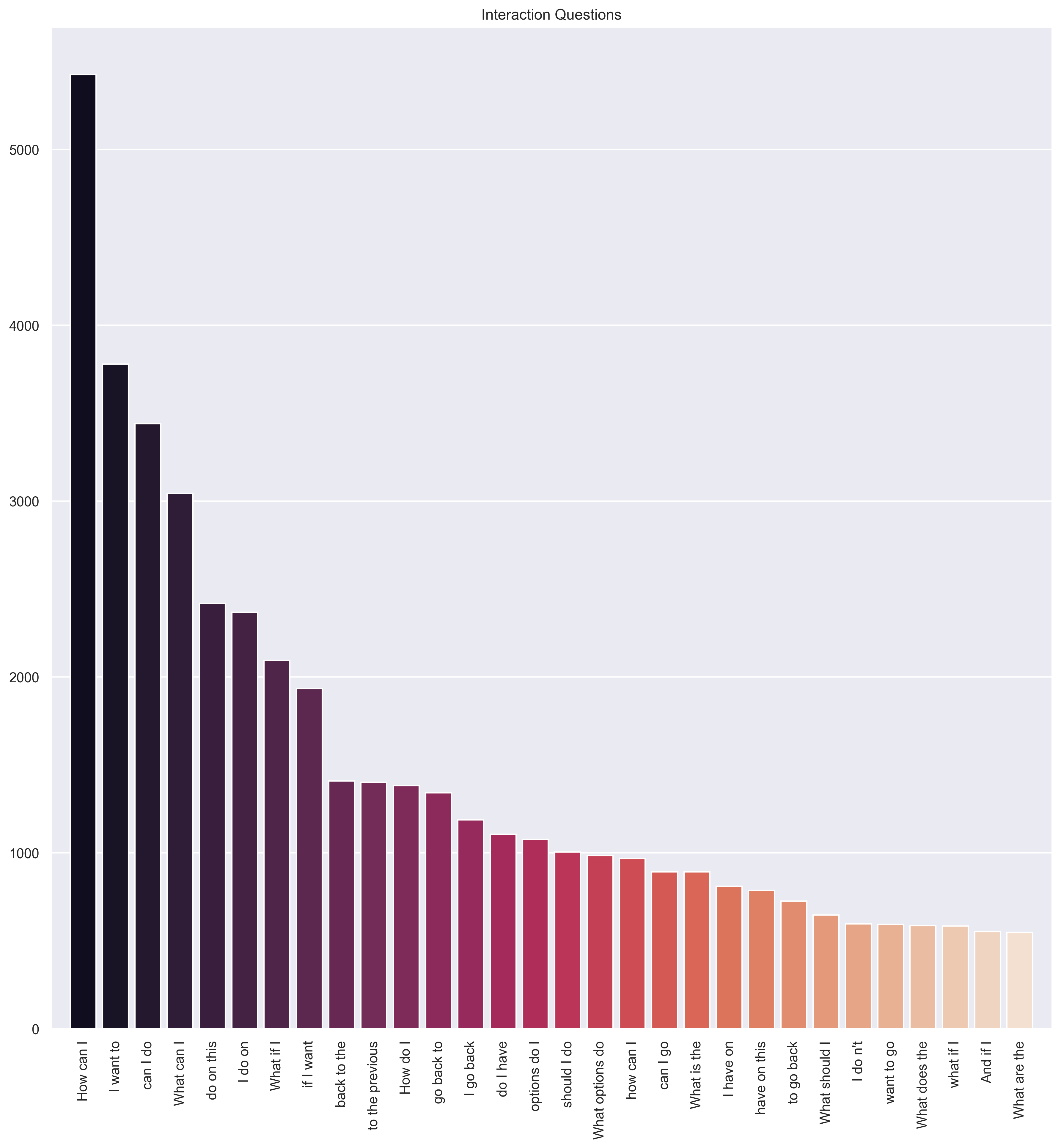}
     \caption{\textit{Conversation interaction} \textbf{questions} trigrams distribution.}
     \label{fig:interaction_q}
 \end{subfigure}
 \hfill
 \begin{subfigure}{0.49\textwidth}
     \includegraphics[width=\textwidth]{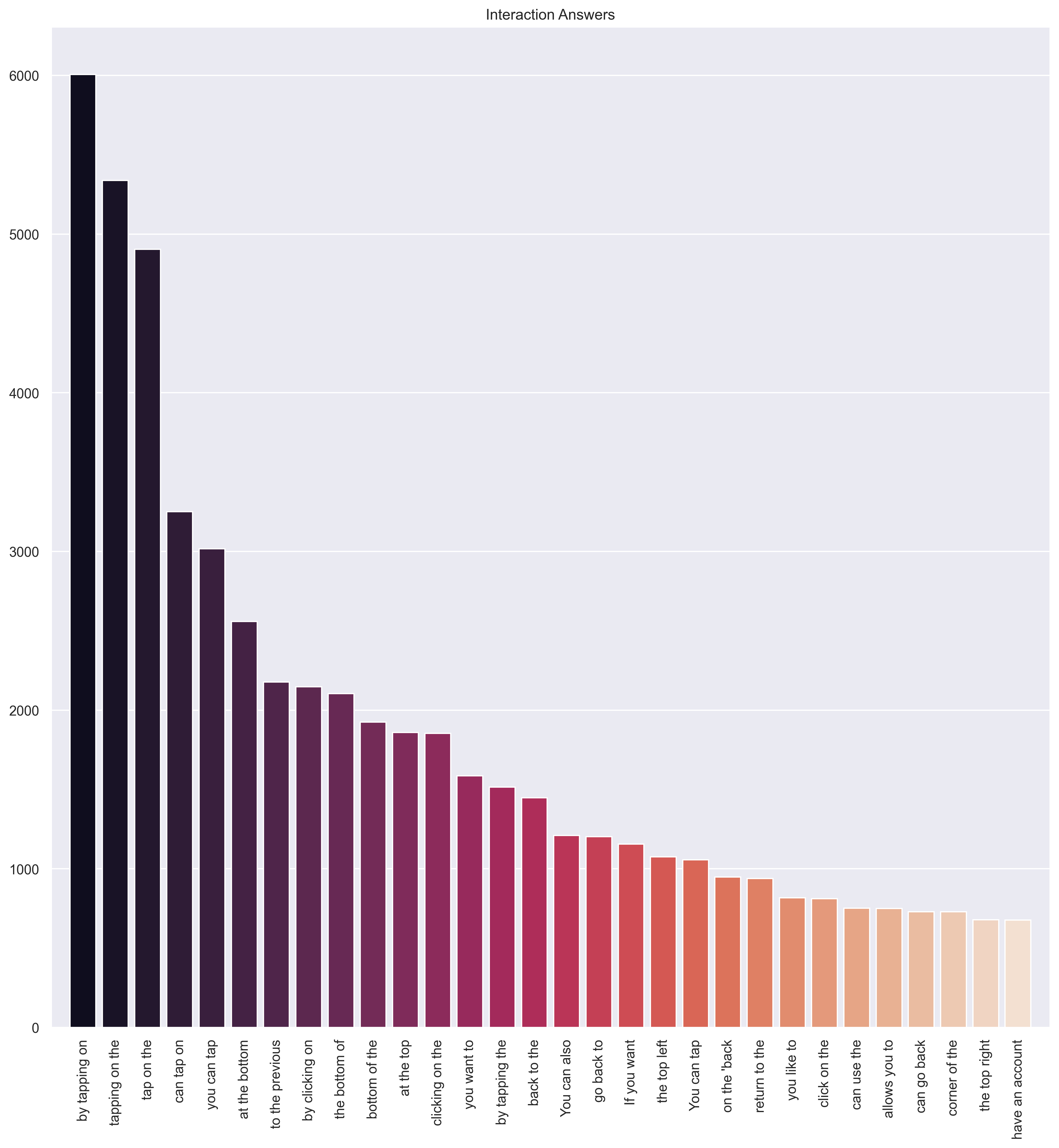}
     \caption{\emph{Conversation interaction} \textbf{answers} trigrams distribution.}
     \label{fig:interaction_a}
 \end{subfigure}

 \caption{Trigrams for collected conversation data questions and answers.}
 \label{fig:conv_data_stat}
\end{figure}

We present detailed distributions of tri-grams in conversation data questions and answers in Fig. \ref{fig:conv_data_stat}. This observation is consistent with our intended objectives for each conversation category, with perception focusing on visual elements and interaction emphasizing actions. Notably, from the interaction conversation answers, we observe that \emph{tap} emerges as the predominant action. In future work, we aim to explore interactions involving other actions, such as scrolling, long-clicking, and entering text. The inclusion of two conversation categories aims to diversify conversation topics, although a clear-cut distinction between the two is not always feasible, and overlap between the categories may occur.

\section{Taperception Label Analysis} \label{appendix:taperception_analysis}

We meticulously label 30 test samples for \emph{taperception} and conduct a study on the correlation among our labels, \emph{taperception} ground-truth labels, Ferret-UI outputs, and GPT-4V outputs. Among the 30 samples, 5 pose challenges in deciphering without direct interaction with the screen.

\begin{figure}[t!]
     \centering
     \begin{subfigure}[b]{0.49\textwidth}
         \centering
         \includegraphics[width=\textwidth]{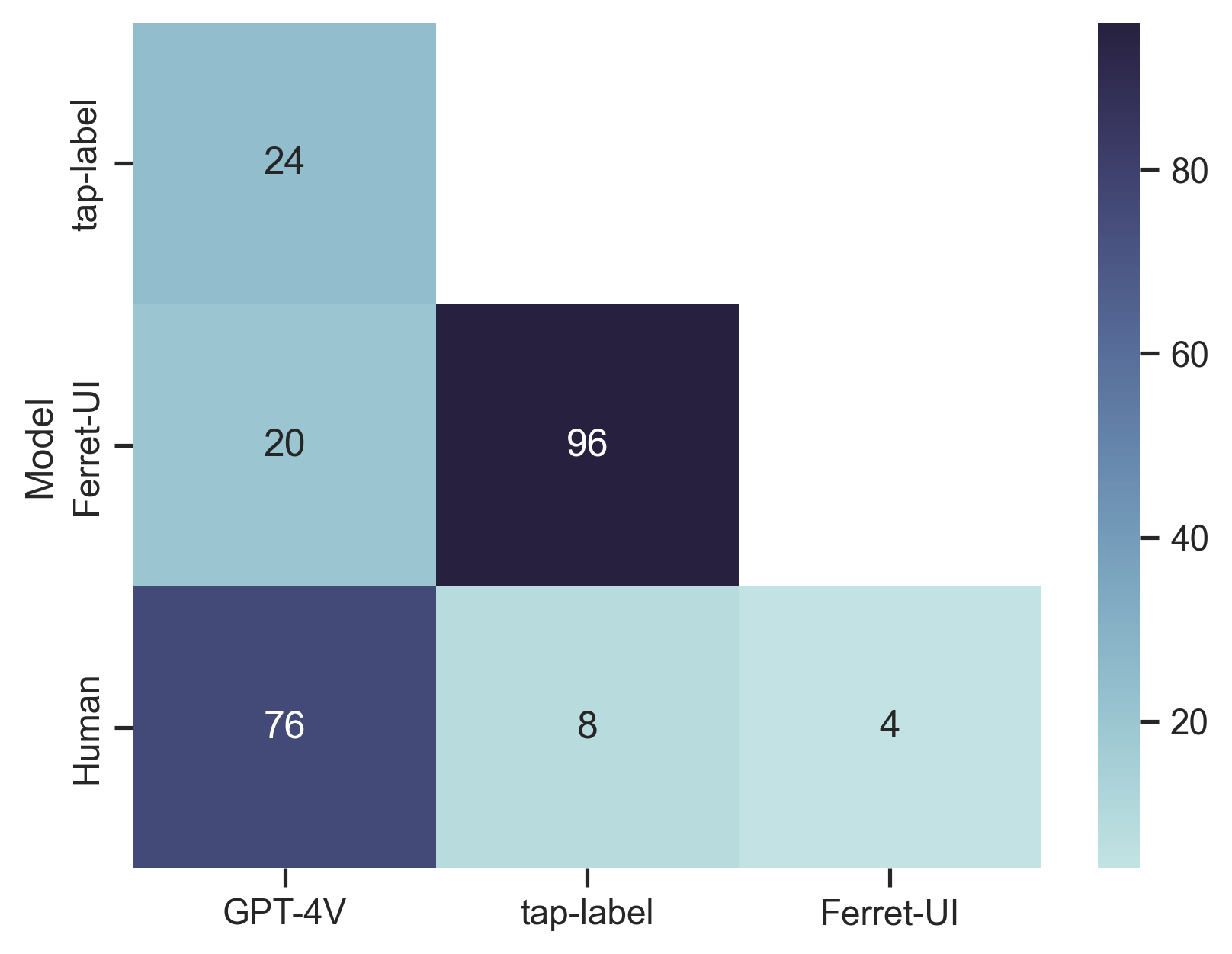}
         \caption{Filtered.}
         \label{fig:y equals x}
     \end{subfigure}
     \begin{subfigure}[b]{0.49\textwidth}
         \centering
         \includegraphics[width=\textwidth]{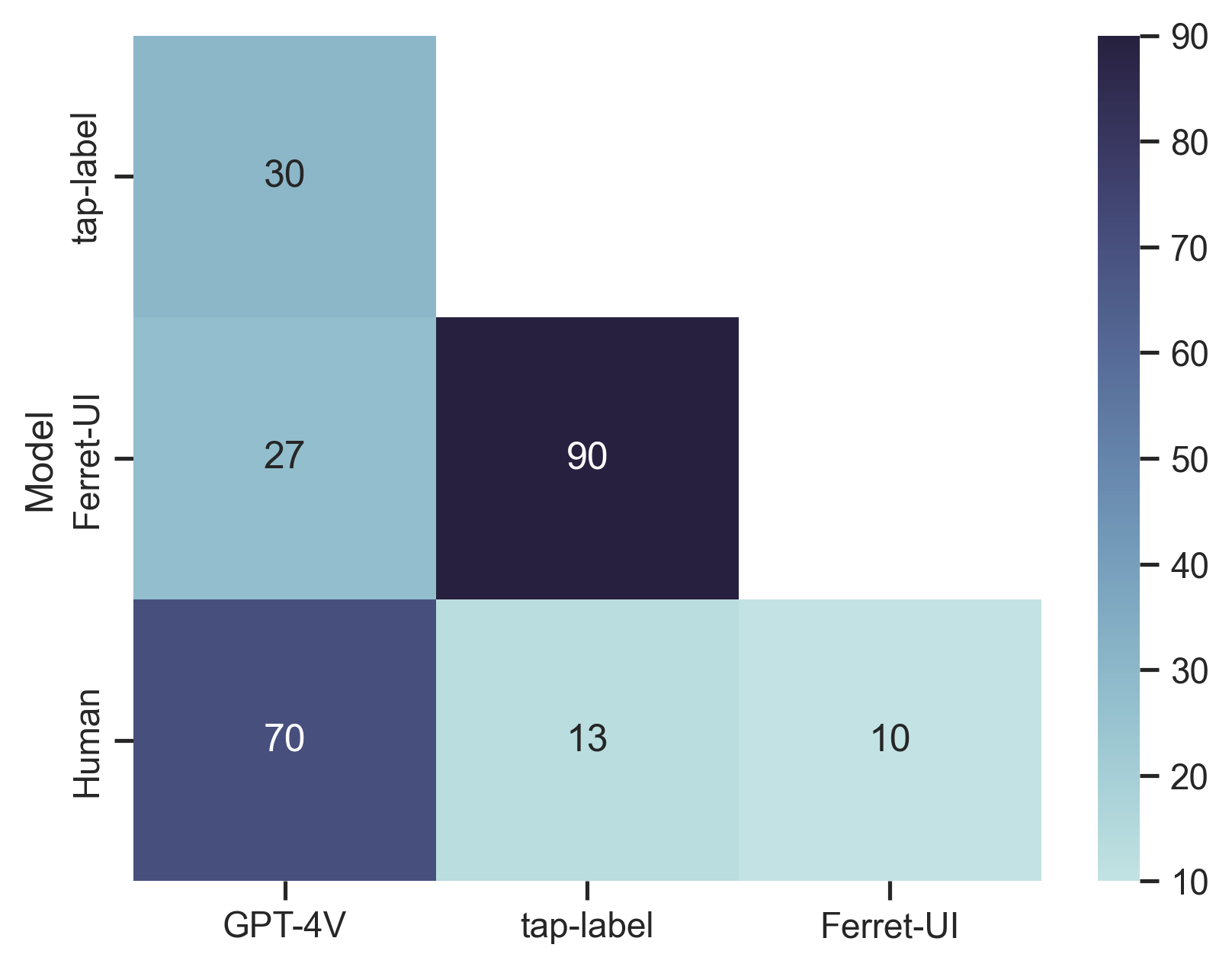}
         \caption{Unfiltered.}
         \label{fig:three sin x}
     \end{subfigure}
        \caption{Agreement between different sources of taperception predictions and labels. In unfiltered, we make the best educational guess for the one that are ambiguous. We observe that our human annotation correlate with GPT-4V (\%76) far more than with taperception label (\%8). Even though Ferret-UI' performance on taperception falls behind compared to Spotlight, it could be due to the noisiness of labels.}
        \label{fig:tap_label_analysis}
\end{figure}

In Tab. \ref{fig:tap_label_analysis}, we present the percentage of agreement among different sources of predictions and labels. The term ``filtered'' denotes the set of 25 instances that are unambiguous, while ``unfiltered'' encompasses the entire 30 instances. Our labels exhibit a high correlation with GPT-4V predictions, but differing significantly from the \emph{taperception} dataset labels. This discrepancy underscores the complexity of predicting \emph{tappability} solely based on single images, highlighting the inherent challenges in obtaining clear-cut labels for this task.

\section{Advanced Task Generation Prompts} \label{appendix:gpt4v_prompts}
\begin{figure}[h!]
    \centerline{
    \includegraphics[scale=0.5]{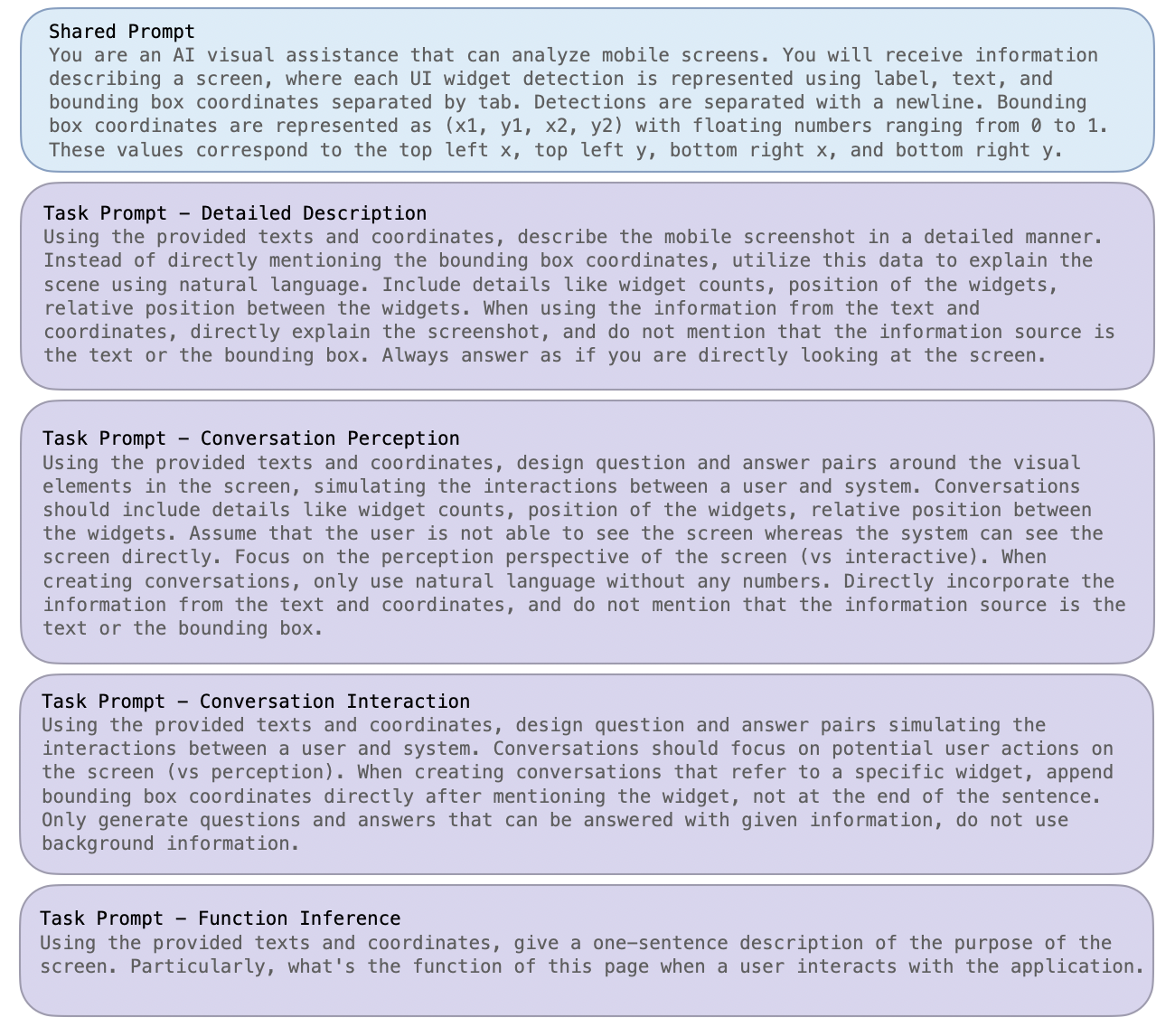}
    }
    \caption{Prompts for GPT-4 in advanced task data generation.} 
    \label{gpt4_prompts}
\end{figure}

We present the prompts to collect advanced task data from GPT-4 in Fig. \ref{gpt4_prompts}.

\begin{figure}[t!]
    \centerline{
        \includegraphics[scale=0.5]{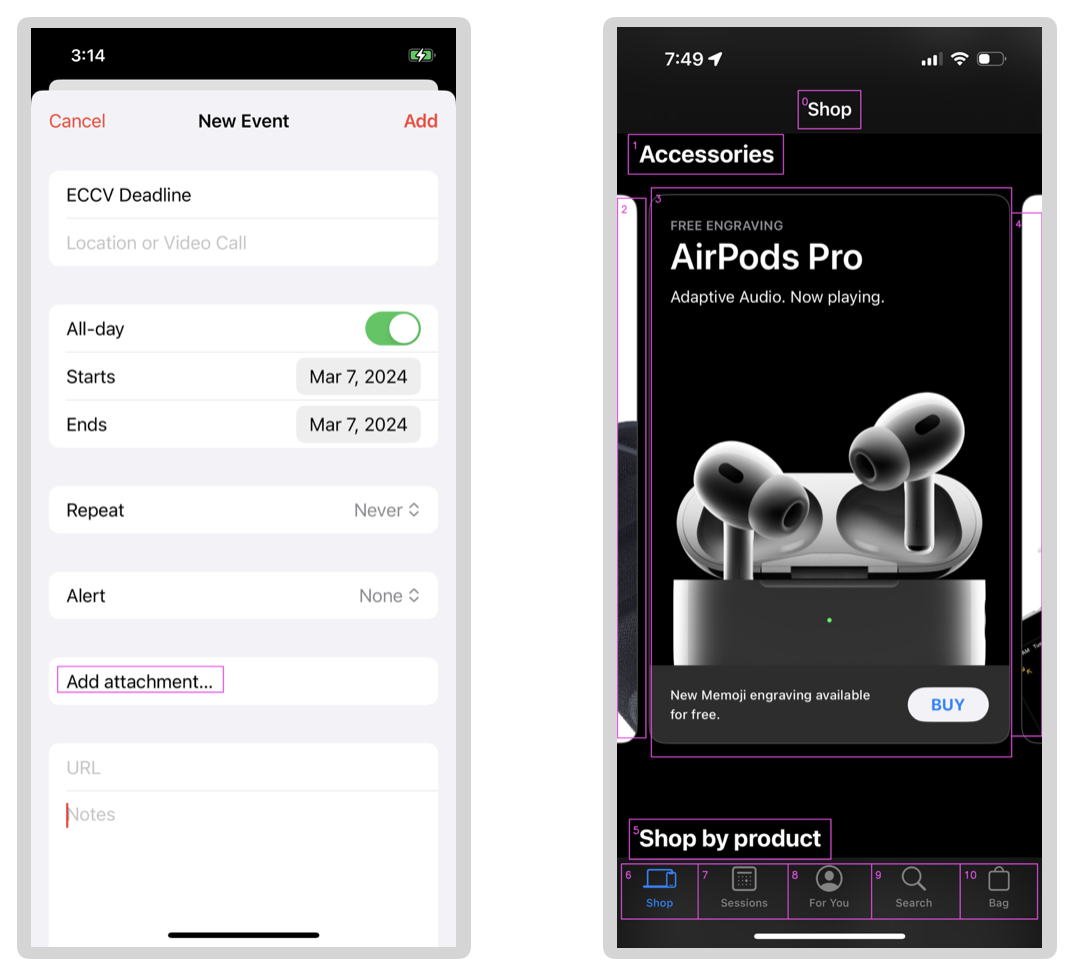}
    }
    \caption{GPT-4V input image examples. Left: used in referring task, where the question concerns one specific UI element. Right: used in grounding task, where GPT-4V refers to the UI elements by their assigned numeric labels.} 
    \label{fig:gpt4v_input}
\end{figure}

\section{GPT-4V Evaluation Details} \label{gpt4v_eval}
We detail the process of creating input for GPT-4V to tackle the UI tasks under scope.

\paragraph{[Input Images]}
We first annotate the screenshots tailored to each specific task, ensuring that GPT-4V has sufficient contextual information to answer the questions. For tasks without any bounding boxes in input or output (\emph{screen2words}, \emph{widget captions}, and \emph{Advanced Tasks}), we use the original images as the input. For tasks that refer to \textbf{one} specific UI element using bounding box in the input, we put a magenta-colored bounding box on the image as the input, as shown in Fig. \ref{fig:gpt4v_input} left. For tasks that expect one or more bounding boxes in the output, our initial explorations confirm that GPT-4V is not able to provide bounding boxes in the output as it gives the answer, \textit{"Unfortunately, I'm not able to provide the exact bounding box coordinates, as my capabilities are currently limited to describing images and discussing the content rather than interacting directly with the image to extract detailed metadata such as pixel coordinates.")} and proceed to answer the question in natural language. Therefore, for those tasks, we create an easier version where we ask GPT-4V to choose from a fixed set of candidates. Particularly, we follow Set-of-Mark prompting~\cite{yang2023set} where for each UI detection from our UI detection model, we use a magenta-colored bounding box to mark it in the screen and inside each box we assign a numeric label so that GPT4-V can refer to it. An example input image is shown in Fig. \ref{fig:gpt4v_input} right.

\paragraph{[Prompts]}
With the input images ready, we further modify the prompts to provide GPT-4V with all the necessary information to perform all the tasks successfully. For taperception, we instruct it to answer \emph{``Yes.''} or \emph{``No.''} only without any explanations. For widget captions, we instruct it to \emph{``Answer in a few words.''} For \emph{icon recognition} and \emph{widget classification}, we provide the list of all possible classes, and instruct it to output the class only without any explanations. For \emph{OCR}, we instruct it to output the identified text only. For \emph{find widget}, \emph{find text}, \emph{find icons}, we add to the prompt \emph{``Candidate choices are indicated using magenta bounding boxes in the image and each box is associated with a numeric label. Output the numeric label as your answer, no need to explain."} 

\section{More Example Outputs}
\begin{figure}[t!]
     \centering
     \begin{subfigure}[b]{1\textwidth}
         \centering
         \includegraphics[width=\textwidth]{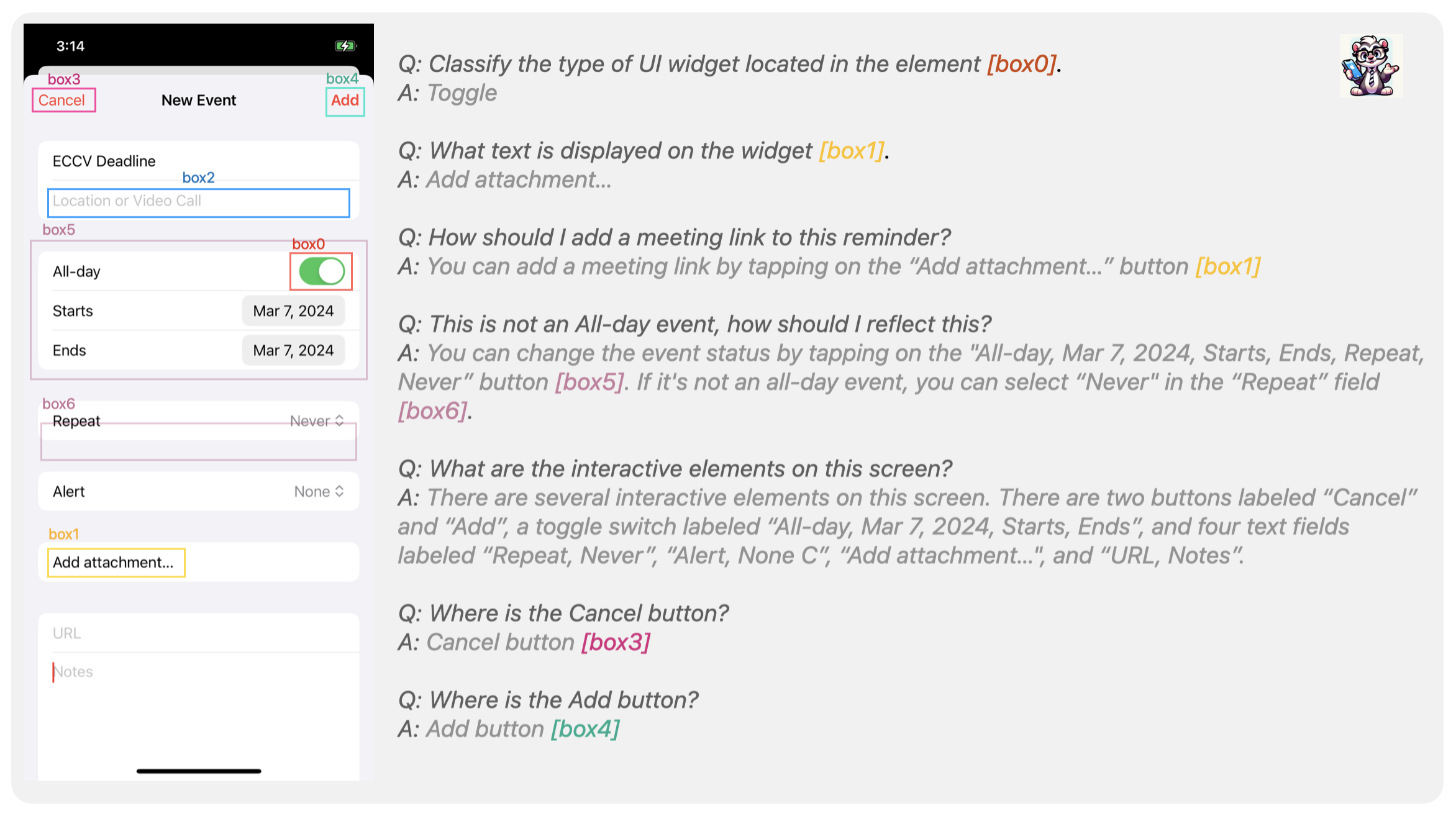}
     \end{subfigure} \\
     \begin{subfigure}[b]{1\textwidth}
         \centering
         \includegraphics[width=\textwidth]{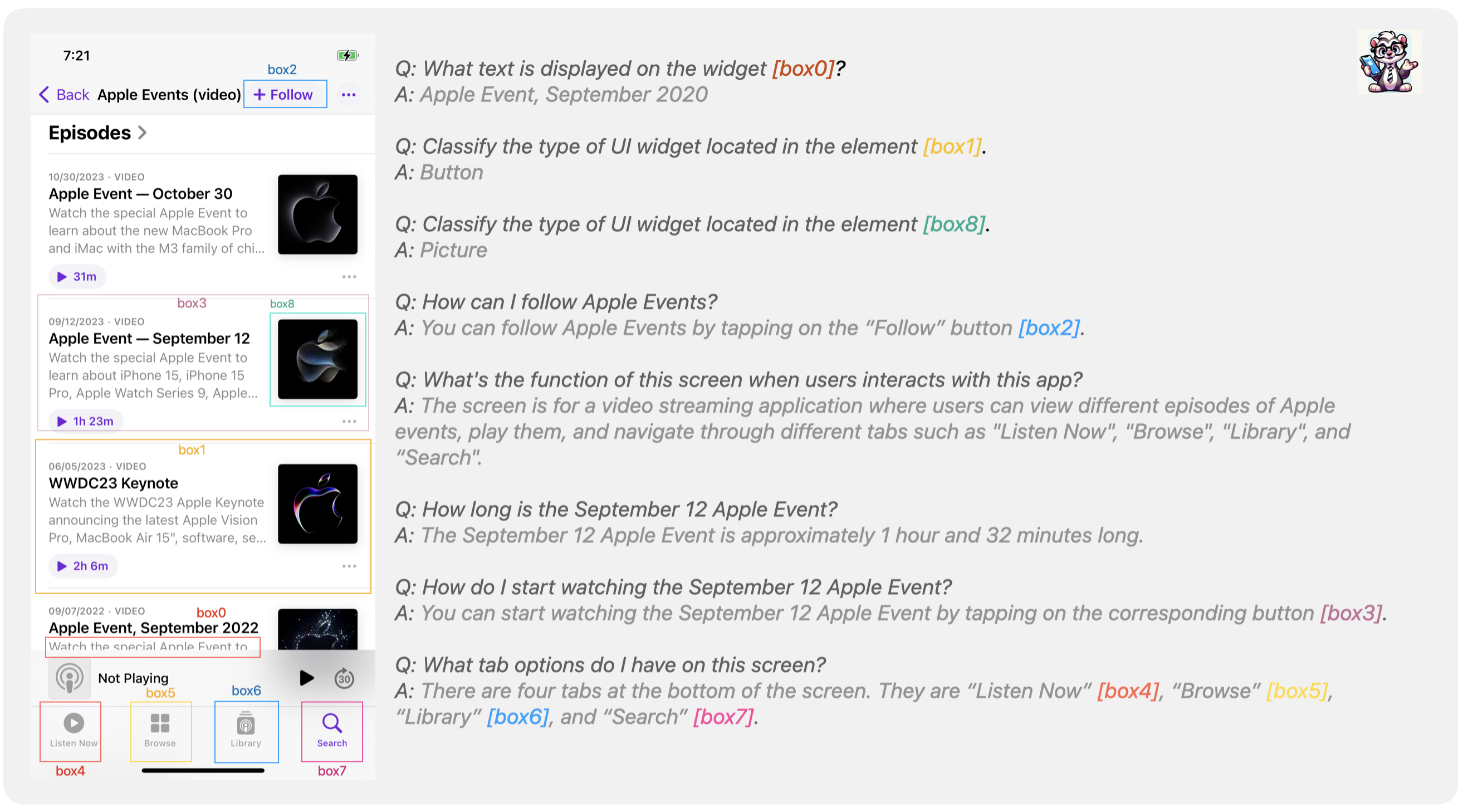}
     \end{subfigure} \\
     \begin{subfigure}[b]{1\textwidth}
         \centering
         \includegraphics[width=\textwidth]{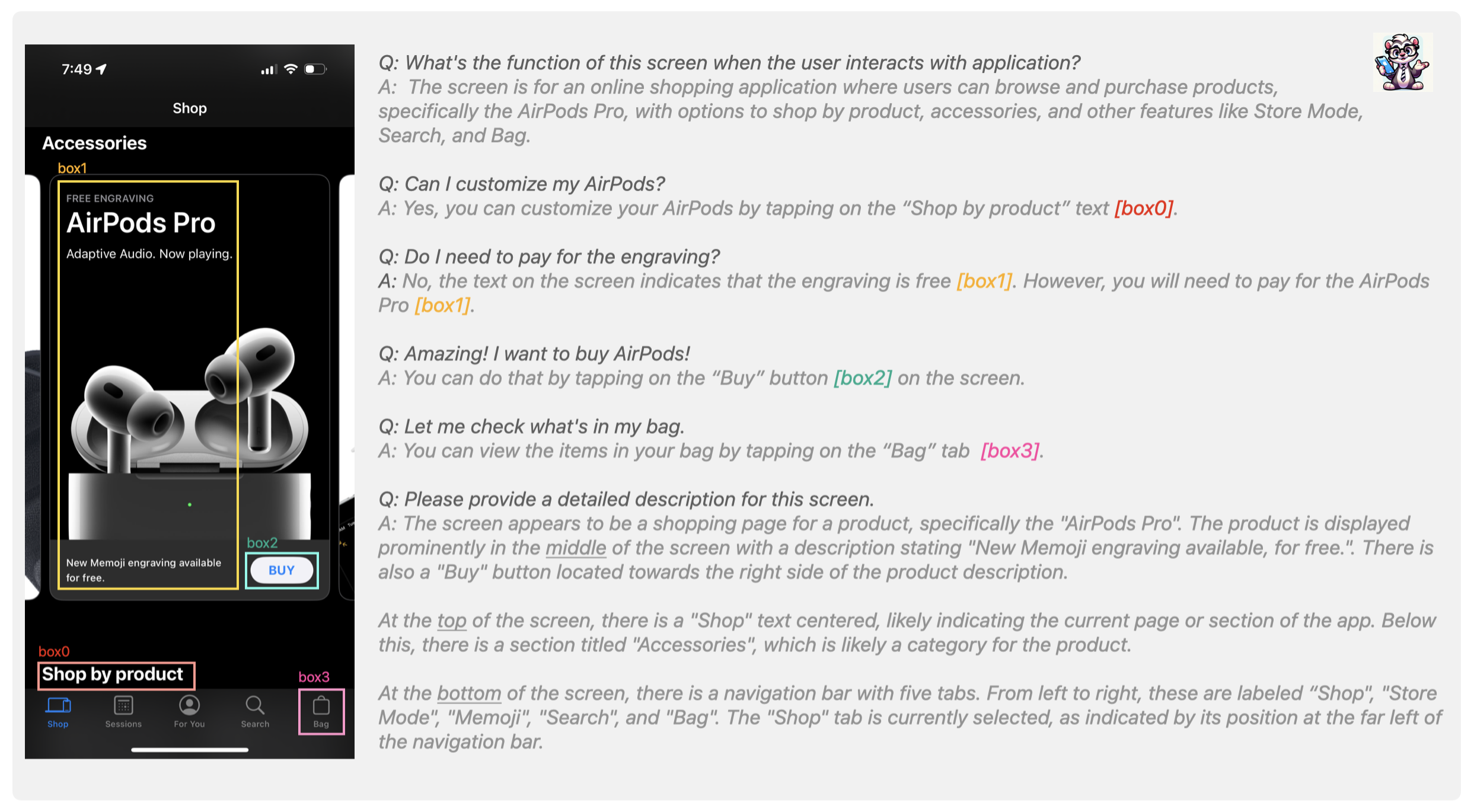}
     \end{subfigure}
        \label{fig:ferret-ui-ex2-3}
\end{figure}

\begin{figure}[t!]
     \centering
     \begin{subfigure}[b]{1\textwidth}
         \centering
         \includegraphics[width=\textwidth]{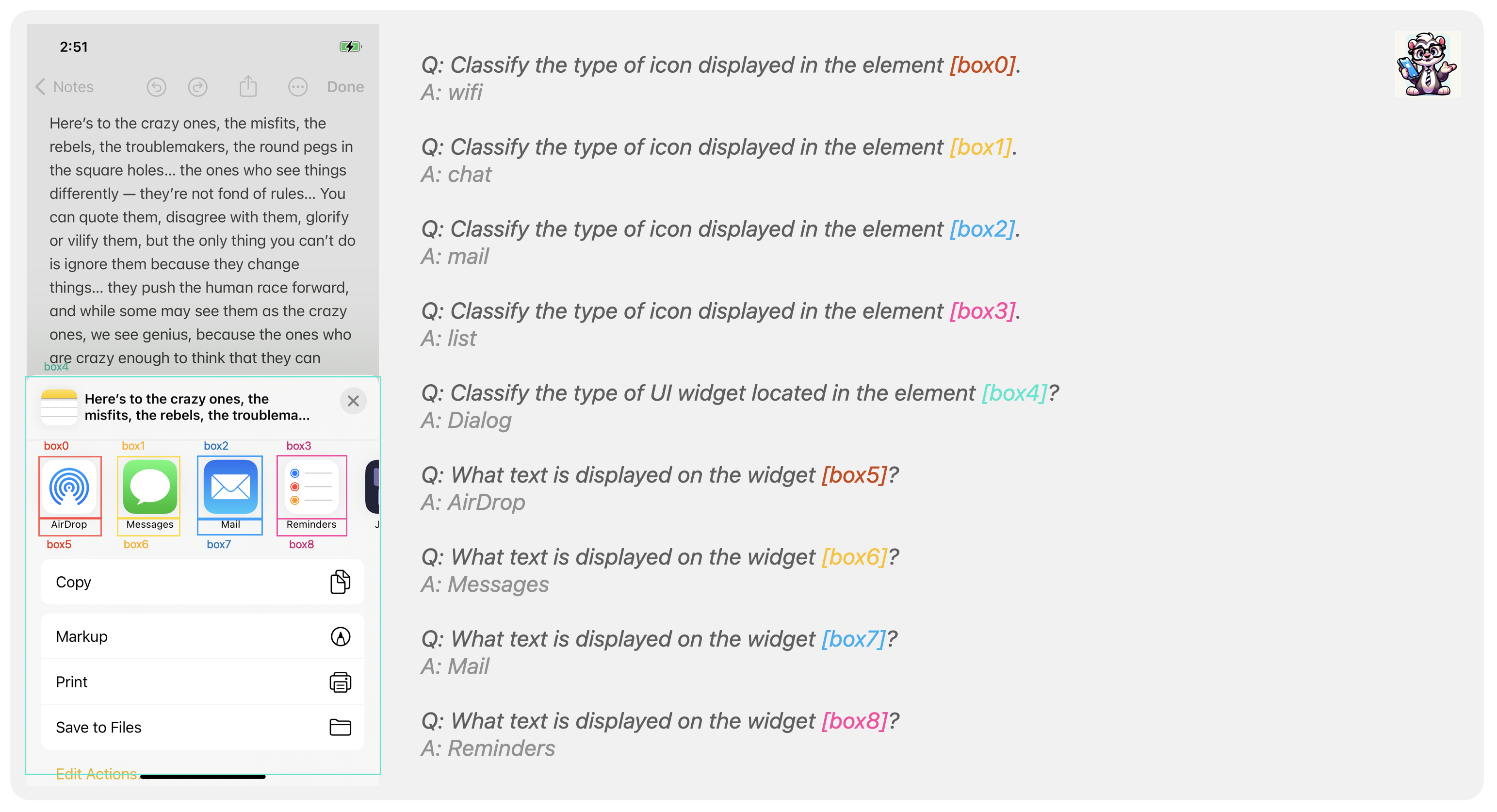}
     \end{subfigure} \\
     \begin{subfigure}[b]{1\textwidth}
         \centering
         \includegraphics[width=\textwidth]{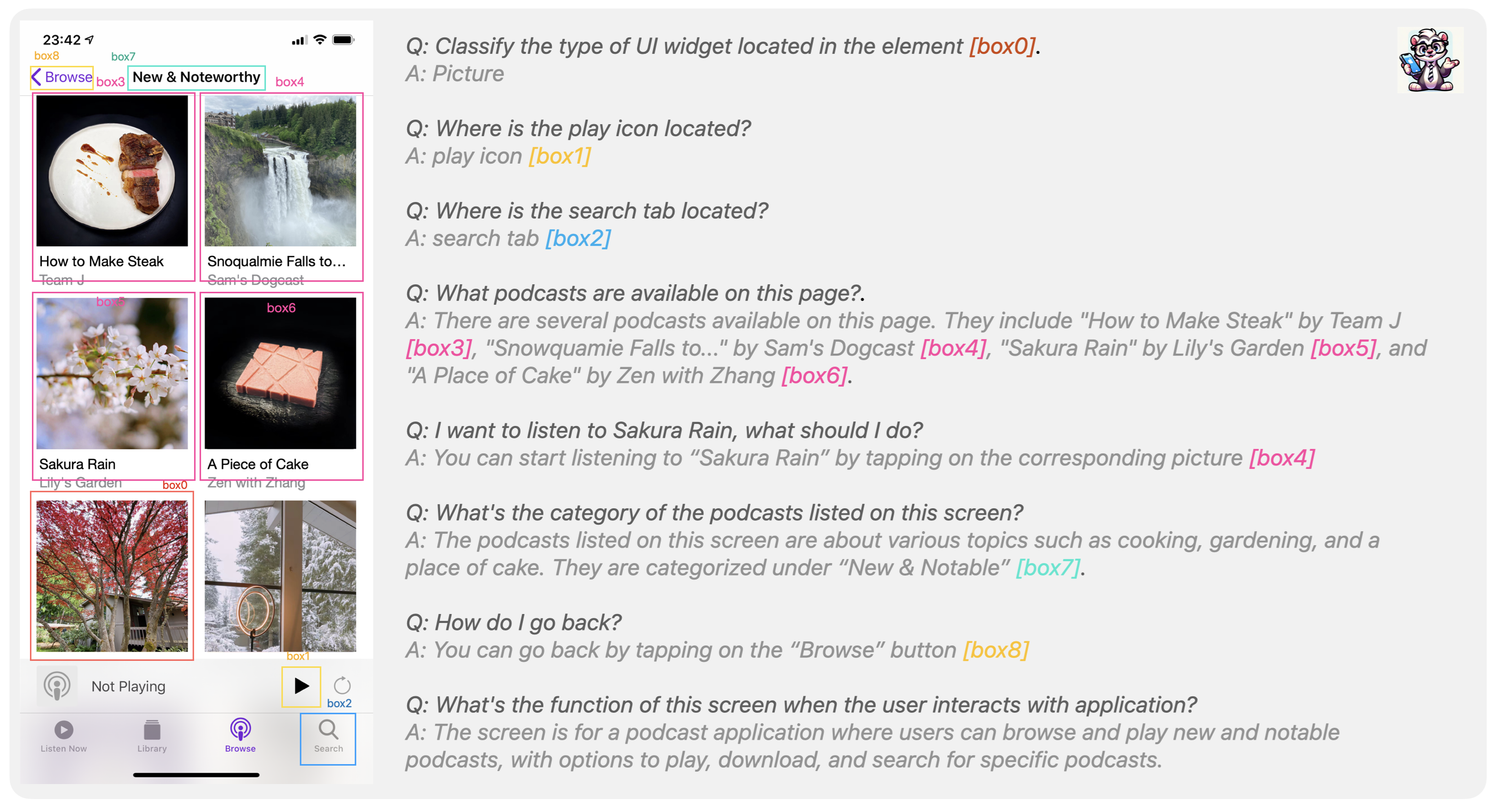}
     \end{subfigure}
        \label{fig:ferret-ui-ex2-3}
\end{figure}

\end{document}